\documentclass[11pt, a4paper, logo, copyright, nonumbering]{klear}
\usepackage{dblfloatfix}
\usepackage{caption}
\usepackage{dramatist}
\usepackage{xspace}
\usepackage{pifont} 
\usepackage{multirow}
\usepackage{tcolorbox}
\usepackage{xltabular}
\usepackage{longtable}
\usepackage{hyperref}
\usepackage{amsfonts}
\usepackage{amsmath}
\usepackage{amssymb}
\usepackage{lineno}
\usepackage{multirow}
\usepackage{adjustbox}

\usepackage[bottom]{footmisc}

\usepackage{CJKutf8}
\usepackage{subfigure}
\usepackage{setspace}

\usepackage{dsfont}
\usepackage{array} %
\usepackage{tabularx} %
\usepackage{subfigure} %
\usepackage{xcolor} %
\usepackage{algorithm}
\usepackage{algorithmic}

\usepackage{lipsum}  %
\usepackage{multicol} %

\usepackage{graphicx}
\usepackage{booktabs}
\usepackage{multirow}

\definecolor{keywordcolor}{rgb}{0.7, 0.1, 0.1}   %
\definecolor{tacticcolor}{rgb}{0.0, 0.1, 0.6}    %
\definecolor{commentcolor}{rgb}{0.4, 0.4, 0.4}   %
\definecolor{symbolcolor}{rgb}{0.0, 0.1, 0.6}    %
\definecolor{sortcolor}{rgb}{0.1, 0.5, 0.1}      %
\definecolor{attributecolor}{rgb}{0.7, 0.1, 0.1} %

\usepackage{listings}

\usepackage{textcomp}
\usepackage[numbers]{natbib}

\definecolor{lightblue}{RGB}{221,235,247}

\definecolor{NavyBlue}{rgb}{0.1, 0.4, 0.8}
\hypersetup{colorlinks = true, linkcolor = NavyBlue,
            urlcolor  = gray,
            citecolor = NavyBlue,
            anchorcolor = NavyBlue}

\usepackage{tcolorbox}
\tcbuselibrary{skins,breakable,xparse} 
\newtcolorbox{promptbox}[1][]{
    colback=gray!10,      %
    colframe=black!50,     %
    boxrule=0.5mm,        %
    arc=1mm,              %
    boxsep=0mm,
    fontupper=\ttfamily\scriptsize,  %
    width=\textwidth,     %
    title=#1,
fonttitle=\ttfamily\footnotesize\centering,
}

\makeatletter
\def\@BTrule[#1]{%
  \ifx\longtable\undefined
    \let\@BTswitch\@BTnormal
  \else\ifx\hline\LT@hline
    \nobreak
    \let\@BTswitch\@BLTrule
  \else
     \let\@BTswitch\@BTnormal
  \fi\fi
  \global\@thisrulewidth=#1\relax
  \ifnum\@thisruleclass=\tw@\vskip\@aboverulesep\else
  \ifnum\@lastruleclass=\z@\vskip\@aboverulesep\else
  \ifnum\@lastruleclass=\@ne\vskip\doublerulesep\fi\fi\fi
  \@BTswitch}
\makeatother

\addto\extrasenglish{
}

 {\begin{list}{}%
         {\setlength{\leftmargin}{#1}}%
         \item[]%
 }
 {\end{list}}

\reportnumber{001} 

\renewcommand{\today}{}

{\centering
\title{AR-GRPO: Training Autoregressive Image Generation Models via Reinforcement Learning}}

\author[*]{
\quad \quad \quad \quad Shihao Yuan, Yahui Liu$^{\heartsuit}$, Yang Yue, Jingyuan Zhang, \newline
Wangmeng Zuo$^{\heartsuit}$, Qi Wang, Fuzheng Zhang, Guorui Zhou
\\
Klear Team, Kuaishou Technology
}

\correspondingauthor{Corresponding author.}

\newcommand\our{LlamaGen}

\begin{document}
\begin{CJK*}{UTF8}{gbsn}

\begin{abstract}
Inspired by the success of reinforcement learning (RL) in refining large language models (LLMs), we propose AR-GRPO, an approach to integrate online RL training into autoregressive (AR) image generation models. 
We adapt the Group Relative Policy Optimization (GRPO) algorithm to refine the vanilla autoregressive models' outputs by carefully designed reward functions that evaluate generated images across multiple quality dimensions, including perceptual quality, realism, and semantic fidelity. 
We conduct comprehensive experiments on both class-conditional (\textit{i.e.}, \textit{class-to-image}) and text-conditional (\textit{i.e.}, \textit{text-to-image}) image generation tasks, demonstrating that our RL-enhanced framework significantly improves both the image quality and human preference of generated images compared to the standard AR baselines.
Our results show consistent improvements across various evaluation metrics, establishing the viability of RL-based optimization for AR image generation and opening new avenues for controllable and high-quality image synthesis.
The source codes and models are available at: \url{https://github.com/Kwai-Klear/AR-GRPO}. 
\end{abstract}

\maketitle

\begin{figure}[h]
\vspace{+1em}
\centering
\includegraphics[width=\linewidth]{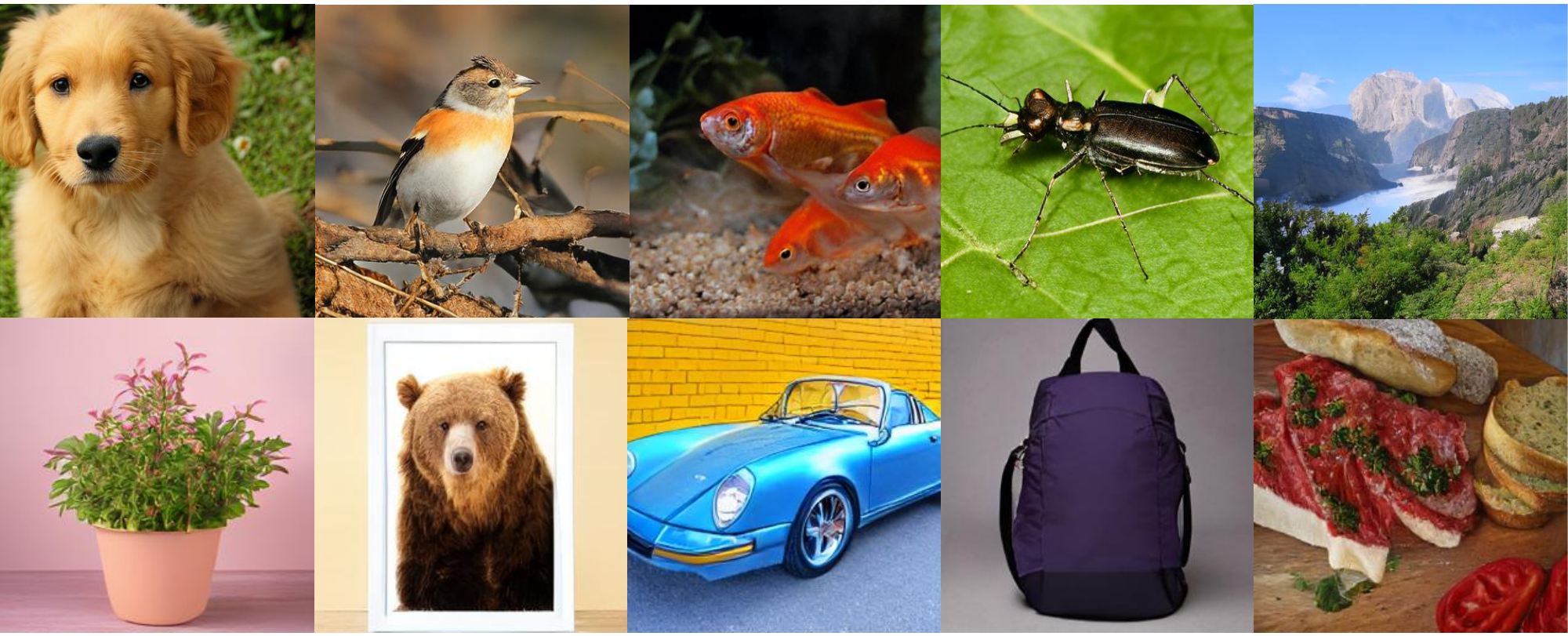}
\caption{Autoregressive image generation enhanced by reinforcement learning based on LlamaGen~\cite{sun2024autoregressive}. We show samples from our
class-conditional image (top row) and text-conditional image (bottom row) generation models.}
\label{fig:teaser}
\end{figure}

\section{Introduction}
\label{sec:introduction}
The field of image generation has witnessed unprecedented progress in recent years, with autoregressive (AR) models~\cite{van2016pixel,van2017neural,chen2020generative} emerging as a powerful paradigm for sequential image synthesis.
Unlike diffusion models~\cite{song2019generative,ho2020denoising,sohl2015deep,song2020score} or GANs~\cite{goodfellow2014generative,zhu2017unpaired,choi2018stargan,karras2020analyzing} that generate images holistically, AR models treat image generation as a sequential prediction problem, generating images pixel by pixel or token by token. This approach has shown particular promise due to its conceptual simplicity and the ability to leverage architectural advances from natural language processing (NLP) field.

The success of reinforcement learning (RL) in optimizing large language models (LLMs), particularly through techniques like reinforcement learning from human feedback (RLHF)~\cite{ouyang2022training} or verifiable feedback (RLVF)~\cite{shao2024deepseekmath,yu2025dapo}, have demonstrated the potential of RL methods to align model output with desired objectives beyond simple likelihood maximization. In the context of language models, RL has proven effective in improving text quality, reducing harmful output, and enhancing task-specific performance. This success naturally raises the question: \textit{can similar RL techniques be applied to improve autoregressive image generation models?}

Traditional training of AR image generation models~\cite{esser2021taming,chen2018pixelsnail,yu2021vector,ramesh2021zero,sun2024autoregressive,tian2024visual} are based primarily on maximum likelihood estimation (MLE), optimizing models to predict the next token given previous tokens. Although this approach has yielded impressive results, it suffers from several limitations. First, MLE optimization may not align directly with the perceptual quality metrics that humans use to evaluate images. Second, MLE training may not effectively encourage image quality in generated output, potentially leading to moderate image quality but lacking fine-grained high-frequency details. Third, the exposure bias problem, where models are trained on ground-truth sequences but generate based on their own predictions during inference, can lead to error accumulation (\textit{i.e.}, artifacts).

Our work addresses these limitations by introducing an RL framework specifically designed for AR image generation models. We apply Group Relative Policy Optimization (GRPO)~\cite{shao2024deepseekmath,guo2025deepseek}, an advanced policy optimization algorithm, to fine-tune pre-trained AR image models. Our main contributions are:
\begin{itemize}
    \item We present the first comprehensive application of GRPO to \textit{autoregressive} image generation, demonstrating how RL techniques from the NLP field can be successfully adapted to the AR image generation tasks. (\textsection\ref{sec:grpo})
    \item We develop sophisticated reward functions that capture various aspects of text-image alignment, image quality, and perceptual realism, enabling fine-grained control over generation objectives for AR image generation models. (\textsection\ref{sec:reward})
    \item  We conduct extensive experiments across two challenging scenarios, including both class-conditional and text-conditional image generation, demonstrating consistent improvements in both image quality and human preference metrics. (\textsection\ref{sec:experiments})
\end{itemize}

Our comprehensive experimental results demonstrate that RL training applied to autoregressive image generation models yields varying impacts across image quality, diversity, and human preference metrics. Below, we summarize our key findings.
\begin{tcolorbox}[colback=lightblue!80,breakable]
\begin{enumerate}[leftmargin=1em]
    \item In class-conditional image generation scenarios, GRPO's encouragement of higher rewards for intra-group sampled images leads to a decline in policy entropy, resulting in more deterministic sampling behavior. This suggests a fundamental trade-off: image quality improvements come at the expense of generation diversity. (\textsection\ref{subsec:class-to-image})
    \item In text-conditional image generation tasks, reward design enables explicit improvements across multiple dimensions—text-image alignment, image quality, and human preference. As expected, GRPO algorithm delivers significant gains across these metrics on various benchmarks, highlighting the substantial potential of RL training for tasks with clearly defined objectives. (\textsection\ref{subsec:text-to-image}) 
    \item We observe a key advantage of RL training in AR image generation: its ability to generalize effectively across both model size and image resolution scaling, highlighting considerable potential for practical deployment. (\textsection\ref{subsec:class-to-image})
\end{enumerate}
\end{tcolorbox}

The remainder of this paper is organized as follows: Section~\ref{sec:related_work} reviews related work in AR image generation and RL applications. Section~\ref{sec:method} details our proposed methodology, including the GRPO algorithm adaptation and reward function design. Section~\ref{sec:experiments} presents comprehensive experimental results on both class-conditional and text-conditional image generation tasks, and detailed analysis and discussion of our findings. Finally, Section~\ref{sec:conclusion} concludes with current limitations and future research directions.

\section{Related Work}
\label{sec:related_work}

\noindent\textbf{Autoregressive Image Generation.} Inspired by the remarkable scalability of autoregressive models in large language models (LLMs), pioneering works have explored applying autoregressive approaches to image generation. Early efforts include VQVAE~\cite{van2017neural,razavi2019generating} and VQGAN~\cite{esser2021taming,lee2022autoregressive}, which introduced the concept of image tokenization by converting continuous images into discrete tokens through learned codebooks. Building upon these foundations, DALL-E~\cite{ramesh2021zero} and Parti~\cite{yu2021vector,yu2022scaling} demonstrated the potential of autoregressive models for text-to-image generation by applying next-token prediction strategies to generate image tokens sequentially.
These early autoregressive approaches showed competitive performance compared to their contemporaries, including GANs~\cite{goodfellow2014generative,brock2018large,karras2020analyzing} and early diffusion models~\cite{song2019generative,song2020denoising,ho2020denoising,dhariwal2021diffusion,rombach2022high,wu2025diffusionreward}, particularly in the pre-2022 era. However, the landscape has evolved significantly with recent developments in autoregressive visual generation. Modern approaches~\cite{sun2024autoregressive,tian2024visual,ControlAR} have addressed many limitations of earlier methods through improved tokenization schemes, better architectural designs, and enhanced training strategies. These advances have renewed interest in autoregressive models as a viable alternative to diffusion-based approaches, particularly given their natural compatibility with language model architectures and training paradigms~\cite{ding2021cogview,alayrac2022flamingo,wu2024janus,ma2025janusflow,chen2025janus}.

\noindent\textbf{Reinforcement Learning Applications.} Recent large language models (LLMs), such as Open AI O1/O3 and Deepseek R1, which are enhanced by posttraining scaling, emerge with numerous powerful and intriguing reasoning behaviors~\citep{guo2025deepseek,claude2025sonnet,qwq32b}. By integrating reinforcement learning from human feedback (RLHF)~\cite{ouyang2022training} or verifiable feedback (RLVF)~\cite{shao2024deepseekmath,yu2025dapo} during post-training, such LLMs have demonstrated impressive performance in solving mathematical, coding, and agent problems through natural language interactions. In the vision and multimodal domains, recent explorations have emerged using reinforcement learning methods to enhance model image generation quality. For example, Flow-GRPO~\cite{liu2025flow} and DanceGRPO~\cite{xue2025dancegrpo} explore the application of RL algorithms to improve Stable Diffusion models. T2I-R1~\cite{jiang2025t2i} proposed a reasoning-enhanced text-to-image generation model powered by RL with a bi-level chain-of-thought (CoT) reasoning process. Up till now, no methods have directly incorporated RL into autoregressive models for image generation, which is the gap this paper aims to address.

\section{Method}
\label{sec:method}

\subsection{Autoregressive Image Generation Framework}
\label{sec:ar-image-gen}
Our approach builds upon autoregressive (AR) image generation models that treat images as sequences of discrete tokens. For brevity, we omit pixel sequence modeling methods~\cite{van2016pixel} here. Given an image $I$, we first tokenize it into a sequence of tokens $\mathbf{x} = (x_1, x_2, ..., x_T)$ using a learned tokenizer~\cite{van2017neural,esser2021taming,yu2021vector}. The AR model then learns to predict the probability distribution over the next token given the previous tokens:
\begin{equation}
\label{eq:ar_base}
    p(x) = \prod_{i=1}^N p(x_t | x_{<t}, c; \theta),
\end{equation}
where $p_(x_t | x_{<t}, c; \theta)$ represents the probability of the current element $x_t$ conditioned on all previous elements $x_{<t}$ in the sequence, with $\theta$ denoting the model parameters and $c$ denotes the conditioning information (class label or text prompt). During training, the model is typically optimized using maximum likelihood estimation (MLE):

\begin{equation}
    \mathcal{L}_\text{MLE} = -\mathbb{E}_{(\mathbf{x}, c)\sim\mathcal{D}} \left[\sum_{t=1}^T\log P(x_t|x_{<t}, c)\right],
\end{equation}
where $\mathcal{D}$ represents the training dataset.

We start with LlamaGen~\cite{sun2024autoregressive}, a Transformer~\cite{vaswani2017attention}-based AR model for image generation. LlamaGen follows the encoder-quantizer-decoder architecture as VQGAN~\cite{esser2021taming}. The encoder and the decoder are convolutional networks with the same downsample ratio. The quantizer contains a codebook $Z\in\mathbb{R}^{K\times C}$ with $K$ learnable vectors. In our experiments, we use the default configurations (\textit{i.e.}, $C=8$ and $K=16384$). Given the tokenized visual tokens, LlamaGen employs the Llama~\cite{llama1} architecture to model token sequences, incorporating pre-normalization with RMSNorm~\cite{zhang2019root}, SwiGLU~\cite{shazeer2020glu} activation functions, and rotary positional embeddings~\cite{su2024roformer}. As shown in Table~\ref{tab:model_scaling}, We employ two models of different scales to validate our proposed RL optimization approach.

\begin{table}[t]
\centering
\begin{tabular}{@{}lcccc@{}}
\toprule
\bf Model & \bf Parameters & \bf Layers & \bf Hidden Size & \bf Heads \\
\midrule
\our-B & 111M & 12 & 768 &  12 \\
\our-L & 343M & 24 & 1024 & 16 \\
\our-XL & 775M & 36 & 1280 &  20 \\
\bottomrule
\end{tabular}
\caption{The three selected baseline models have different sizes and architecture configurations from \our~\cite{sun2024autoregressive}. The configurations are following previous works~\cite{gpt2,llama1}.
}
\label{tab:model_scaling}
\end{table}

\subsection{Policy Optimization Algorithm}
\label{sec:grpo}

We formulate the AR image generation task as a Markov Decision Process (MDP) where:
\begin{itemize}
    \item \textit{State Space} $\mathcal{S}$: The state $s_t\in\mathcal{S}$ at the time step $t$ consists of the conditioning information $c$ and the sequence of previously generated tokens $x_{<t}$.
    \item \textit{Action Space} $\mathcal{A}$: The action $a_t\in\mathcal{A}$ corresponds to selecting the next token $x_t$ from the vocabulary (\textit{i.e.}, a learned codebook).
    \item \textit{Policy}: The autoregressive model $\pi_\theta(a_t | s_t) = P(x_t | x_{<t}, c)$ serves as our policy, which is parameterized by $\theta$.
    \item \textit{Reward Function}: We design reward functions $r(\mathbf{x}, c)$ that evaluate the quality of the complete generated sequence. Considering that evaluating individual tokens is often meaningless, we only evaluate the complete predicted sequence $\mathbf{x}$ in conjunction with the input condition $c$.
\end{itemize}

We employ Group Relative Policy Optimization (GRPO)~\cite{shao2024deepseekmath} as our primary RL algorithm to optimize the MDP. 
For each input condition $c$, 
GRPO samples a group of outputs $\{o_1, o_2, \cdots, o_G\}$ from the old policy $\pi_{\theta_{old}}$ and then collect the feedback for the group of responses through our reward system. According to each feedback, we assign a particular reward. Then, the advantage of the $i$-th output is calculated by normalizing the group-level rewards $\{R_1, R_2, \cdots, R_G\}$:
\begin{equation}
\hat{A}_{i,t} = \frac{r_i - \text{mean}(\{R_i\}_{i=1}^G)}{\text{std}(\{R_i\}_{i=1}^G)}.
\end{equation}
Finally, we optimizes the policy model $\pi_{\theta}$ by maximizing the following objective:
\begin{equation}
\begin{aligned}
\mathcal{J}_\text{GRPO}(\theta)& = \mathbb{E}_{(\mathbf{x},c)\sim \mathcal{D}, \{o_i\}_{i=1}^G\sim \pi_{\theta_\text{old}}(\cdot\mid c)} \\&
\Bigg[ \frac{1}{G}\sum_{i=1}^{G} 
\frac{1}{|o_i|}
\sum_{t=1}^{|o_i|} \Bigg( 
\min \Big( r_{i,t}(\theta) \hat{A}_{i,t},  
\ \text{clip} \Big( r_{i,t}(\theta), 1 - \varepsilon, 1 + \varepsilon \Big) \hat{A}_{i,t} \Big)
- \beta D_{\text{KL}}(\pi_{\theta} || \pi_{\text{ref}}) 
\Bigg) \Bigg],
\label{eq:grpoloss}
\end{aligned}
\end{equation}
where 
\begin{equation}
    r_{i,t}(\theta)=\frac{\pi_{\theta}(o_{i,t} \mid c, o_{i,<t})}{\pi_{\theta_{\text{old}}}(o_{i,t} \mid c,o_{i,<t})},
\end{equation}
and $\varepsilon$ is a hyperparameter. In our experiments, we set $\varepsilon=0.2$. $D_\text{KL}(\cdot\|\cdot)$ denotes to Kullback–Leibler (KL) divergence that prevents the trained policy from deviating too far from the reference policy. $\beta$ is a hyper-parameter that controls the regularization strength. The KL divergence is estimated with the following unbiased estimator\citep{kl_approx}: 
\begin{equation}
\small
    \mathbb{D}_\text{KL}(\pi_{\theta} || \pi_{ref}) = \frac{\pi_{ref}(o_{i,t}|c,o_{i,<t})}{\pi_{\theta}(o_{i,t}|c,o_{i,<t})}- \log\frac{\pi_{ref}(o_{i,t}|c,o_{i,<t})}{\pi_{\theta}(o_{i,t}|c,o_{i,<t})} - 1,
\end{equation}
which is guaranteed to be positive. 
In the following Section~\ref{sec:reward}, we present the details of our reward design. 

\subsection{Reward Design}
\label{sec:reward}

To accurately evaluate the quality of generated images, we convert the complete predicted sequence $\mathbf{x}$ to the corresponding image $\mathbf{I}$ using the decoder component of the encoder-quantizer-decoder architecture. 
Then, we develop multi-faceted reward functions that capture different aspects of image quality:
\begin{itemize}
    \item \textit{Conditional Reward}: We incorporate CLIP~\cite{radford2021learning} and HPSv2~\cite{wu2023human} to ensure generated images are consistent with the input condition:
    \begin{equation}
        r_\text{C} = \text{CLIP}(\mathbf{I}, c) + \text{HPSv2}(\mathbf{I}, c). 
    \end{equation}
    \item \textit{Image Quality Reward}: We employ a no-reference image quality assessment method, such as MANIQA~\cite{yang2022maniqa}, to ensure generated images are visually appealing:
    \begin{equation}
        r_\text{I} = \text{MANIQA}(\mathbf{I}).
    \end{equation}
    \item \textit{Realism Reward}: Since CLIP-like models primarily focus on the presence of visual semantic information while MANIQA emphasizes detail assessment, neither comprehensively evaluates image realism and both may overlook critical issues such as image artifacts, which are essential considerations for image generation quality. Therefore, we employ an additional vision-language model (VLM) (\textit{i.e.}, Qwen2.5-VL-3B-Instruct~\cite{qwen2.5-VL}) to judge image realism from three aspects (See details in Appendix~\ref{app:vlm-prompts}):
    \begin{equation}
        r_\text{R} = \text{VLM}(p, c, \mathbf{I}),
    \end{equation}
    where $p$ refers to the instruction prompt. 
\end{itemize}
Therefore, our final reward is formulated as:
\begin{equation}
    r_\text{final} = \lambda_\text{C} r_\text{C} + \lambda_{I}r_\text{I} + \lambda_{R}r_\text{R}, %
\end{equation}
where $\lambda_\text{C}$, $\lambda_\text{I}$ and  $\lambda_\text{R}$ are reward weights. In our experiments, we set their values to 1 (See more details in Appendix~\ref{app:reward-details}).

\subsection{Training Procedure}
\label{sec:training}

We fine-tune the pre-trained model (\textit{i.e.}, LlamaGen~\cite{sun2024autoregressive}) using GRPO~\cite{shao2024deepseekmath} algorithm with our designed reward functions. Following LlamaGen, we conduct training on two representative image generation tasks:
\begin{itemize}
    \item \textit{Class-conditional Image Generation.} It is also termed class-to-image (C2I) generation.  The class embedding is indexed from a set of learnable embeddings~\cite{peebles2023scalable,esser2021taming} and is used as the pre-filling token embedding. 
    Starting from this token embedding, the model generates the sequence of image tokens through next-token prediction and stops upon reaching the predefined maximum length.
    \item \textit{Text-conditional Image Generation.} It is also termed text-to-image (T2I) generation. We use FLAN-T5 XL~\cite{chung2024scaling} as the text encoder to integrate the text condition into AR models. The encoded text features are projected through an additional MLP~\cite{chen2024gentron} and serve as prefilling token embeddings in AR models. Similarly, statring from the pre-filling token embeddings, the model generates the sequence of image tokens through next-token prediction and stops upon reaching the maximum length.
\end{itemize}

Additionally, we retain the classifier-free guidance (CFG)~\cite{ho2022classifier} module during RL training, which it is well-known for improving visual quality and text-image alignment. During training, the conditional input is randomly dropped and replaced with a null unconditional embedding~\cite{peebles2023scalable}. In inference, for each token, its logit $\ell_{g}$ is formulated by $\ell_{g} = \ell_u + s(\ell_c - \ell_u)$, where $\ell_c$ and $\ell_u$ are conditional logit and unconditional logit respectively, and $s$ is scale of the classifier-free guidance. 

The RL training alternates between three core phases: (1) \textit{Generation Phase.} Sample multiple sequences (\textit{i.e.}, a group of outputs with size $G$) from the current policy for each conditioning input. (2) \textit{Evaluation Phase.} Compute rewards for all generated samples using our multi-faceted reward function.
(3) \textit{Optimization Phase.} Update policy parameters using GRPO objective with computed advantages.

\section{Experimental Results}
\label{sec:experiments}

\subsection{Class-conditional Image Generation}
\label{subsec:class-to-image}

\noindent\textbf{Training Setups.} 
For class-conditional generation, we continue training  LlamaGen~\cite{sun2024autoregressive} with the same data as it is pretrained on ImageNet~\cite{imagenet}. The image resolution is set $256\times 256$. During the sampling process of GRPO, we use the same configurations as the pretrained model with top\_k = 0~(all), top\_p = 1.0, temperature = 1.0. We set CFG parameter during training to 2.0 for default. Detailed discussion of CFG weight in \textsection\ref{subsec:ablation}. In each step, we sample 8 candidates for each input with the on-policy training strategy. The KL constraint is enabled and $\beta$ is set to $0.1$. To stabilize RL training, all dropout rates in LlamaGen are set to 0. The training parameters are the same for all experiments, with a total batch size of 64. The Optimizer is AdamW~\cite{adamw} with a learning rate of $1e-5$ and default $\beta_1=0.9$, $\beta_2=0.95$, and weight decay $=0.05$. 

\begin{table}[htp]
\centering
\begin{tabular}{@{}lccccccc@{}}
\toprule
\bf Model & \bf Image Size & \bf IS$\uparrow$ & \bf FID$\downarrow$ & \bf sFID$\downarrow$ & \bf Precision$\uparrow$ & \bf Recall$\uparrow$ \\
\midrule
\our-B & \multirow{2}{*}{256} & 193.61 & 5.46 &  7.50 & 0.84 & 0.46 \\ 
 \quad + RL (Ours) & & 211.67 & 5.91 & 8.58 & 0.86 & 0.40  \\ \midrule
\our-B & \multirow{2}{*}{384} &  157.17 & 6.44 &  7.49 &  0.81 & 0.46 \\ 
 \quad + RL (Ours) & & 195.45 & 7.64 & 6.84 & 0.88 & 0.34 \\ \midrule
 \our-L & \multirow{2}{*}{384} & 256.07 & 3.08 & 6.09 & 0.83 & 0.52  \\ 
 \quad + RL (Ours) & & 264.64 & 4.24 & 5.91 & 0.87 & 0.44 \\ \midrule
\our-XL & \multirow{2}{*}{384} &  286.88 & 2.78 &  5.57 &   0.84 &  0.54 \\
 \quad + RL (Ours) &  & 293.07 & 3.63 & 5.99 & 0.86 & 0.48  \\
\bottomrule
\end{tabular}
\\[0.5em]
\caption{Detailed performance on class-conditional ImageNet benchmark. We present comparisons between baseline LlamaGen~\cite{sun2024autoregressive} models of varying sizes and resolutions and their counterparts enhanced with our proposed method. Note that we test all models with the CFG parameter $s = 2.0$. The generated image as $384\times 384$ resolution is resized to $256\times 256$ in the evaluating. 
}
\label{tab:c2i-main-results}
\end{table}

\noindent\textbf{Evaluation Metrics.} 
The evaluation of class-conditional generation is based on ImageNet~\cite{imagenet} with commonly used metrics including Frechet Inception Distance (FID)~\cite{heusel2017gans}, Inception Score (IS)~\cite{salimans2016improved,barratt2018note}, sFID~\cite{nash2021generating}, Precision and Recall~\cite{kynkaanniemi2019improved}.  All evaluations are implemented using ADM’s TensorFlow scripts~\cite{dhariwal2021diffusion} for fair comparisons. 

\noindent\textbf{Results on ImageNet.} As shown in Table~\ref{tab:c2i-main-results}, we compare the baseline LlamaGen models of varying sizes and resolutions and their counterparts enhanced with our proposed RL method. 
We observe that finetuning with RL algorithm enables the model to significantly improve IS and Precision performance while keeping FID and sFID scores comparable.
Comprehensively considering the scores of IS, FID, Precision, and Recall, we find that RL adjusted the distribution of the model's generation results (\textit{i.e.}, slightly weakening FID), allowing the generated results to demonstrate higher quality (\textit{i.e.}, higher IS). At the same time, the visual content of the generated images shows higher accuracy in matching the visual semantics of corresponding categories (\textit{i.e.}, higher Precision), though coverage or diversity is slightly compromised (\textit{i.e.}, lower Recall).
We hypothesize that this could be due to the GRPO optimization algorithm encouraging group sampling generation and favoring results with greater positive rewards, which to some extent promotes the generation of more deterministic visual content. We observe similar results in ``B'', ``L'' and ``XL'' size models with larger output resolution (\textit{i.e.}, $384\times 384$).

When we observe experiments on model and image size scaling, we notice that when the model size is relatively small, increasing the image size increases the difficulty for the model to process image sequences, resulting in degraded model performance (\textit{e.g.},  B-256 $\to$ B-384). In this situation, continuously increasing the model size brings clear and sustained benefits (\textit{e.g.}, B-384 $\to$ L-384 $\to$ XL-384).

\begin{figure}[h]
\centering
\includegraphics[width=0.72\linewidth]{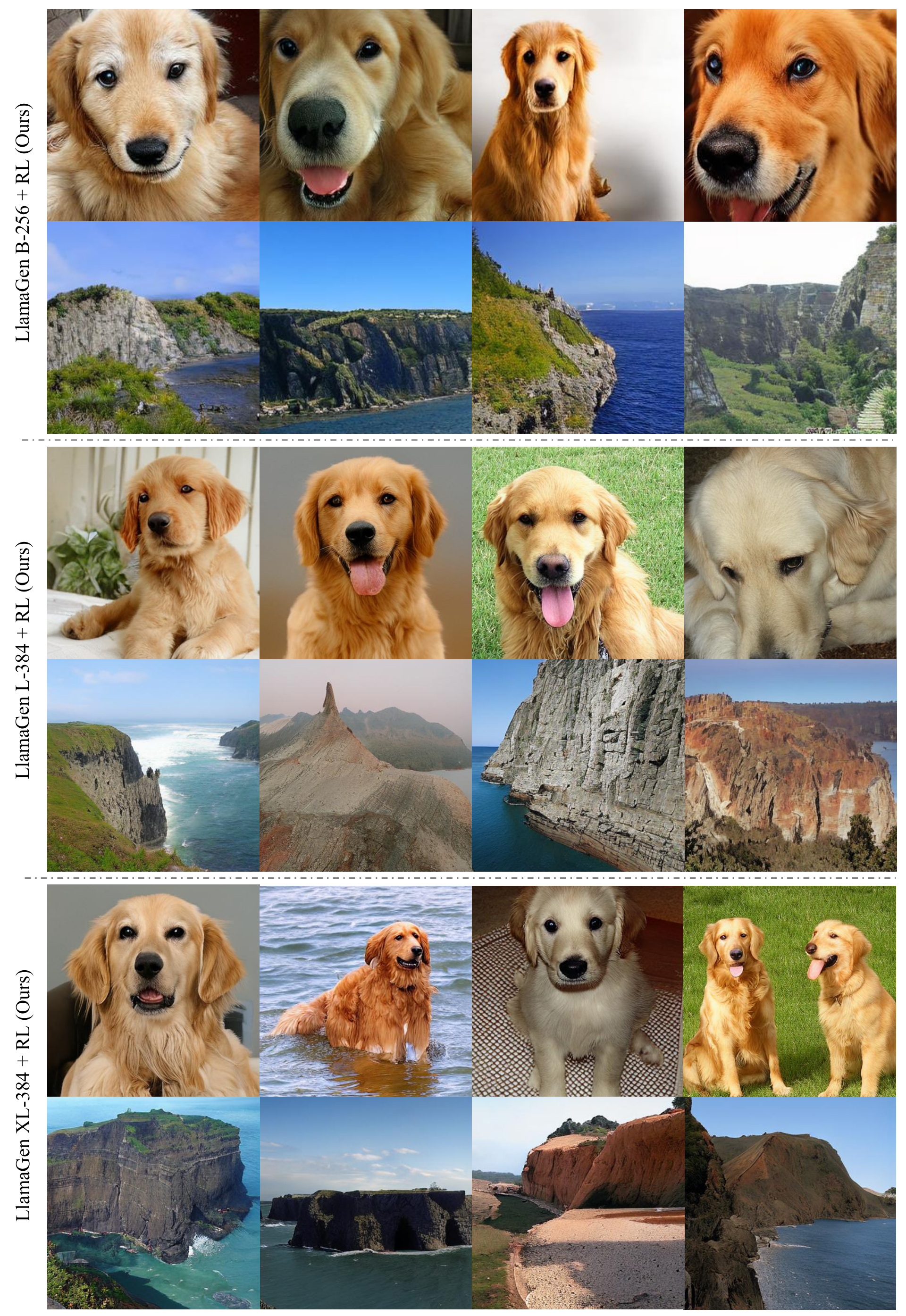}
\caption{Visual results for class-conditional image generation following RL training under various configurations: model sizes ``B'', ``L'', and ``XL'', and resolutions of 256$\times$256 and 384$\times$384.}
\label{fig:c2i-scaling}
\end{figure}

\noindent\textbf{Visual Results.} We present visual results for class-conditional image generation on ImageNet following RL training under various configurations in Figure~\ref{fig:c2i-scaling}.
Our observations reveal that increasing model size and image resolution leads to progressively enhanced image quality, with generated images displaying greater detail and closer resemblance to real-world scenes. This demonstrates the robust generalization capability of RL training across different model size and image resolution scaling regimes. We refer readers to Appendix~\ref{app:visual-results} for more visual results for class-conditional image generation.

\begin{figure}[h]
\centering
\includegraphics[width=0.8\linewidth]{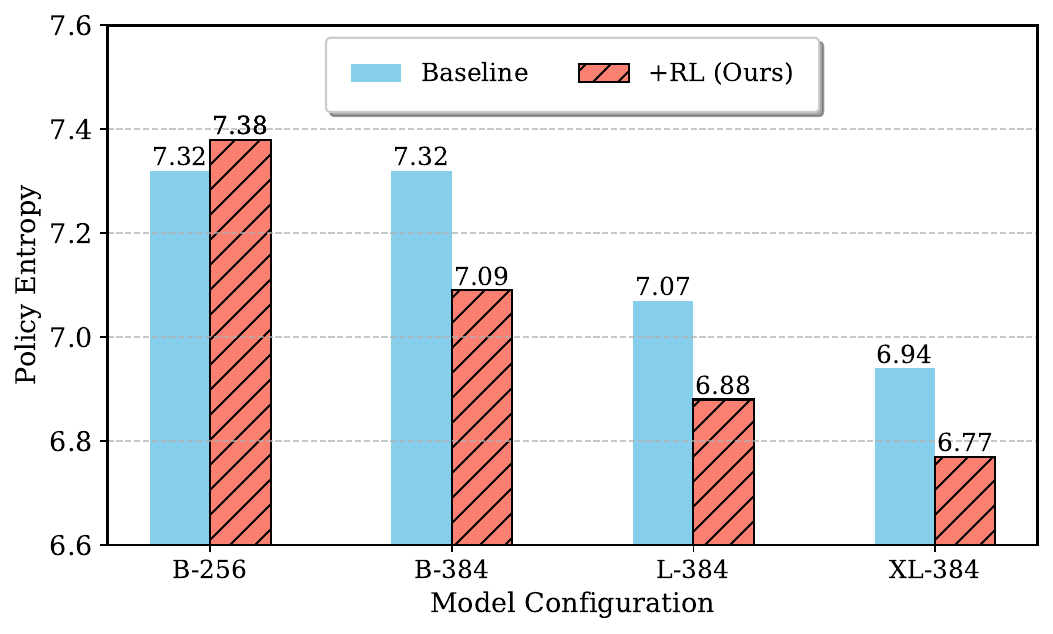}
\caption{Policy entropy comparison between baseline and RL-trained models under various architectural configurations on class-conditional image generation scenarios. ``B'', ''L'' and ``XL'' refer to the model size and ``256'' and ``384'' denote to the image resolution.}
\label{fig:c2i-entropy}
\end{figure}

\noindent\textbf{On Policy Entropy.} Recent work~\cite{cui2025entropy,wang2025beyond,cheng2025reasoning,li2025entropy} has conducted an in-depth analysis of the relationship between the statistical state changes of LLMs' entropy and their exploration capabilities during RL training. We compute the policy entropy of the RL trained models with various configurations on the whole ImageNet validation set. As shown in Figure~\ref{fig:c2i-entropy}, we find that policy entropy exhibits a downward trend, indicating RL-trained models exhibit more deterministic sampling behavior. This aligns with our experimental observations: image quality improvements at the expense of generation diversity.

\subsection{Text-conditional Image Generation}
\label{subsec:text-to-image}

\noindent\textbf{Training Setups.} 
For text-conditinal generation, our RL training is based on the LLamaGen~\cite{sun2024autoregressive} T2I model after stage I with FLAN-T5 XL ~\cite{chung2024scaling} extracting text embeddings. Similar to LlamaGen, we train on the subset of LAION-COCO~\cite{laion-coco} with the image resolution $256\times 256$. Both training and inference CFG is set to  2.0 in text-conditional generation. The other setting is the same as class-conditional generation.

\noindent\textbf{Evaluation Metrics.} GenEval~\cite{ghosh2023geneval} assesses T2I models on complex compositional prompts—like object counting, spatial relations, and attribute binding—across six difficult compositional image generation tasks. We use its official evaluation pipeline, which detects object bounding boxes and colors, then infers their spatial relations. Unlike Flow-GRPO~\cite{liu2025flow}, we do not directly involve GenEval related rewards during our RL training. Furthermore, we incorporate DrawBench~\cite{saharia2022photorealistic}, a comprehensive benchmark consisting of structured text prompts designed for text-to-image model evaluation. We involve more assessment metrics, including text-image alignment (\textit{i.e.}, CLIP score~\cite{radford2021learning} and HPSv2~\cite{wu2023human} score), image qualities (\textit{i.e.},  Aesthetic~\cite{aesthetics} score, DeQA~\cite{you2025teaching}) and human preference alignment (\textit{i.e.},   ImageReward~\cite{xu2023imagereward}, PickScore~\cite{kirstain2023pick}, and UnifiedReward~\cite{wang2025unified}). 

\noindent\textbf{Results on GenEval.} As shown in Table~\ref{tab:t2i-geneval-results}, employing our proposed RL method boosts the performance of base model on \textit{color}, \textit{counting} and \textit{single object}. This is attributed to the limited visual semantics covered by our adopted conditional, image quality and realism rewards. We also observe that recent work Flow-GRPO~\cite{liu2025flow} integrates benchmark-targeted rewards during training, including OCR and GenEval rewards, enabling better performance in specific task scenarios such as position, attribute binding, multiple objects and text rendering. 
Our method can also easily integrate more similar rewards, but this work mainly aims to focus on presenting a general framework for integrating RL fine-tuning into AR image generation tasks. Therefore, this can serve as a promising direction for future exploration.

\begin{table}[h]
\centering
\begin{tabular}{@{}lcccccccc@{}}
\toprule
\bf Model &  \bf PT$\uparrow$ & \bf CL$\uparrow$ & \bf AB$\uparrow$ & \bf CT$\uparrow$ & \bf SO$\uparrow$ & \bf TO$\uparrow$ & \bf Overall \\
\midrule
\our-XL~\cite{sun2024autoregressive}  & 0.042 &  0.550 & 0.032 & 0.197 & 0.750 & 0.263 & 0.306  \\
 \quad + RL (Ours)   & 0.040 & 0.593 & 0.030 & 0.228 & 0.791 & 0.263 & \textbf{0.324} \\ 
\bottomrule
\end{tabular}
\caption{GenEval results. We present comparisons between baseline LlamaGen model and its counterpart enhanced with our proposed method on text-conditional $256\times 256$ image generation. %
``PT'', ``CL'', ``AB'', ``CT'', ``SO'' and ``TO'' refer to \textit{position}, \textit{color}, \textit{attribute binding}, \textit{counting}, \textit{single object} and \textit{two object}, respectively.
}
\label{tab:t2i-geneval-results}
\end{table}

\noindent\textbf{Results on DrawBench.} As shown in Table~\ref{tab:t2i-drawbench-results}, we present results of experiments on text-to-image generation using the DrawBench benchmark, comparing the baseline and our RL trained version. 
The RL enhanced version (``+ RL'') shows consistent improvements across all metrics compared to the baseline model. The results demonstrate that incorporating RL into the text-to-image generation model leads to better alignment with text prompts, higher image quality, and stronger alignment with human preferences.

\begin{table}[h]
\small
\centering
\begin{tabular}{@{}lcccccccc@{}}
\toprule
\bf \multirow{2}{*}{Model} & \multicolumn{2}{c}{\bf Alignment}  & \multicolumn{2}{c}{\textbf{Image Quality}} & \multicolumn{3}{c}{\textbf{Preference Score}} \\ \cmidrule(lr){2-3}\cmidrule(lr){4-5} \cmidrule(lr){6-8}
& \textbf{CLIP Score} & \textbf{HPSv2} &  \textbf{Aesthetic} & \textbf{DeQA}  &\textbf{ImgRwd} & \textbf{PickScore} &  \textbf{UniRwd} \\
\midrule
\our-XL~\cite{sun2024autoregressive} & 0.245 & 0.153 & 4.701 & 0.546  & -1.784 & 0.737 & 0.271 \\
 \quad + RL (Ours) & \textbf{0.274} & \textbf{0.208} & \textbf{4.808} & \textbf{0.551}  &  \textbf{-0.712} & \textbf{0.769} & \textbf{0.312} \\ 
\bottomrule
\end{tabular}
\caption{Text-image alignment, image quality and human preference alignment results on DrawBench~\cite{saharia2022photorealistic} text-to-image $256\times 256$ generation benchmark. Our method shows improvements across all metrics.
}
\label{tab:t2i-drawbench-results}
\end{table}

\begin{figure}[h]
\centering
\includegraphics[width=0.8\linewidth]{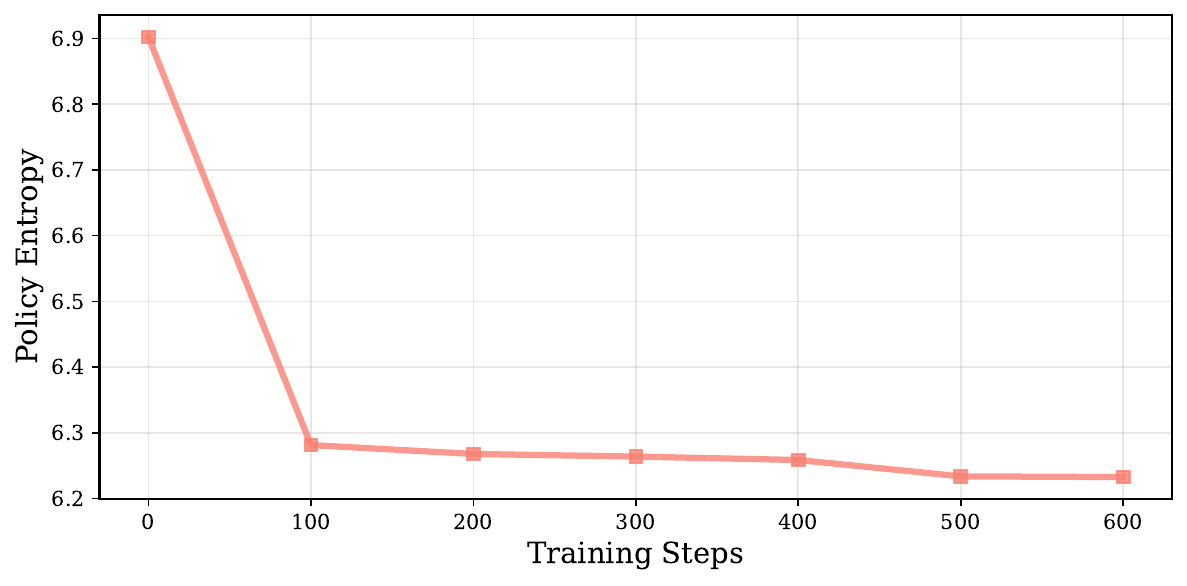}
\caption{Policy entropy curve throughout the training process on text-conditional image generation task.}
\label{fig:t2i-entropy}
\end{figure}

\noindent\textbf{On Policy Entropy.} We compute the policy entropy on GenEval benchmark, as shown in Figure~\ref{fig:t2i-entropy}. We observe a decline trend on the policy entropy between the baseline (\textit{i.e.}, step 0) and the intermediate checkpoints saved during the RL training. This observed pattern is consistent with results obtained in class-conditional image generation scenarios, with both findings indicating that models demonstrate increased sampling determinism after undergoing GRPO training.

\subsection{Ablation Study}
\label{subsec:ablation}
\noindent\textbf{On Effectiveness of KL Penalty.} In Table~\ref{tab:c2i-kl-ablation}, we present an ablation study examining the effect of removing the KL penalty from the GRPO algorithm for image generation at $256\times 256$. The KL penalty serves as a crucial regularization mechanism in RL training for image generation. Without it, the model achieves higher IS and Precision but suffers: (1) \textit{Mode collapse}, evidenced by very low recall (0.32), indicating the model generates high-quality but limited diversity of images; (2) \textit{Distribution drift}, worse FID scores suggest generated images deviate more from the real data distribution. Therefore, the KL penalty helps maintain a better balance between quality and diversity, preventing the model from collapsing to a narrow set of high-quality outputs while preserving reasonable image fidelity. 

\begin{table}[htp]
\centering
\begin{tabular}{@{}lccccccc@{}}
\toprule
\bf Model & \bf Image Size & \bf IS$\uparrow$ & \bf FID$\downarrow$ & \bf sFID$\downarrow$ & \bf Precision$\uparrow$ & \bf Recall$\uparrow$ \\
\midrule
\our-B & \multirow{3}{*}{256} & 193.61 & 5.46 &  7.50 & 0.84 & 0.46 \\ 
 \quad + RL w/o KL & & 220.57 & 7.83 & 11.72 & 0.88 & 0.32  \\ 
 \quad + RL w/ KL & & 211.67 & 5.91 & 8.58 & 0.86 & 0.40  \\ 
\bottomrule
\end{tabular}
\caption{Ablation study on removing KL penalty.}
\label{tab:c2i-kl-ablation}
\end{table}

\begin{figure}[t]
\begin{minipage}{0.49\textwidth}
    \includegraphics[width=\textwidth]{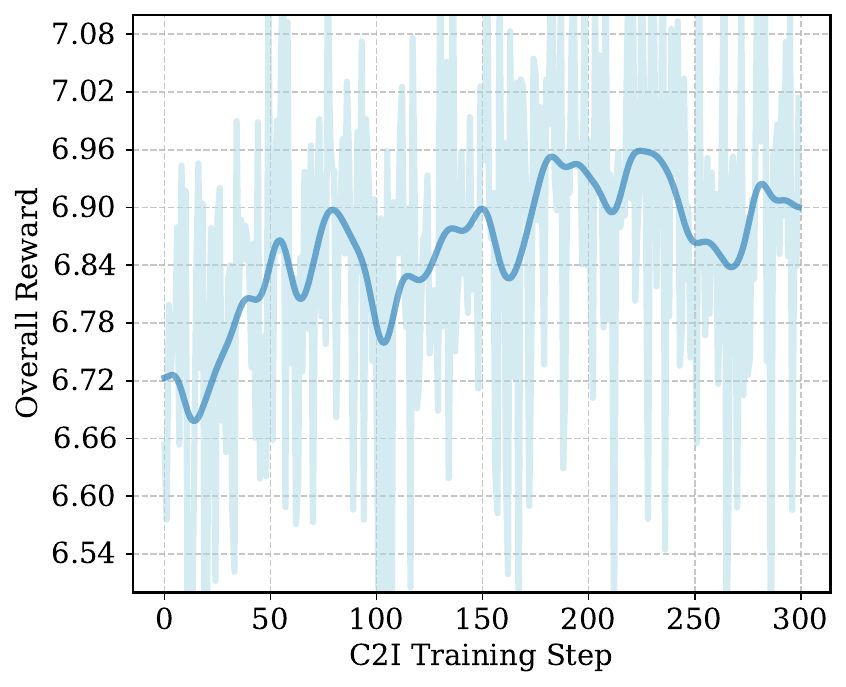}
\end{minipage}
 \hfill
\begin{minipage}{0.49\textwidth}
    \centering
    \includegraphics[width=\textwidth]{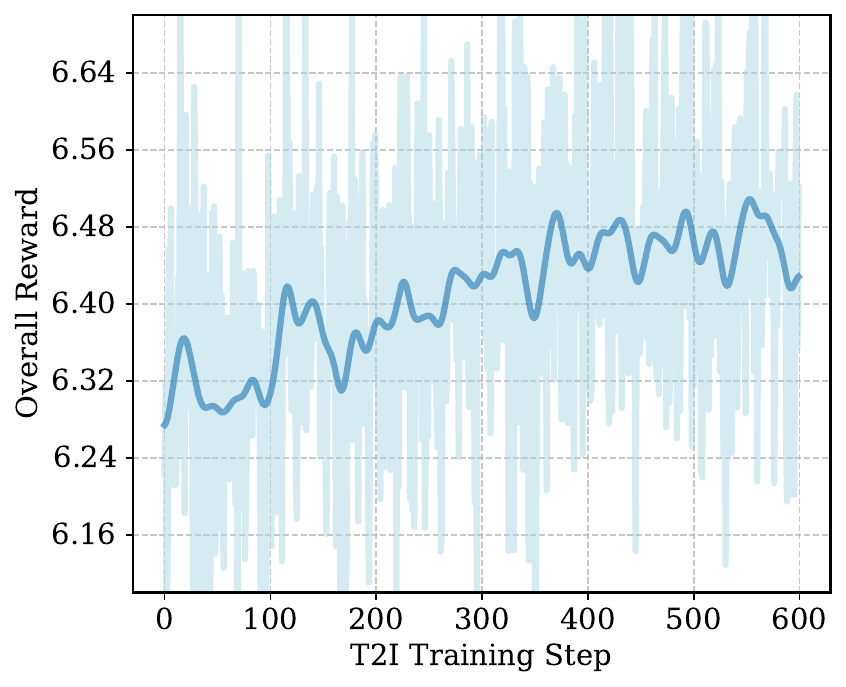}
\end{minipage}
\caption{Overall reward curves as a function of training steps during reinforcement learning for C2I (Left) and T2I (Right) image generation models.}
\label{fig:overall-reward}
\end{figure}

\begin{table}[h]
\centering
\begin{tabular}{@{}lcccccccc@{}}
\toprule
\bf \multirow{2}{*}{Model} &  \multicolumn{3}{c}{\bf Reward Design} & \multirow{2}{*}{\bf IS$\uparrow$} & \multirow{2}{*}{\bf FID$\downarrow$} & \multirow{2}{*}{\bf sFID$\downarrow$} & \multirow{2}{*}{\bf Precision$\uparrow$} & \multirow{2}{*}{\bf Recall$\uparrow$} \\ \cmidrule(lr){2-4}
& $\lambda_\text{C}$ & $\lambda_\text{I}$ & $\lambda_\text{R}$ & & & \\
\midrule
\our-B &  &  &  & 193.61 & 5.46 &  7.50 & 0.84 & 0.46 \\ 
 \quad RL (A): & 0 & 1.0 & 1.0 & 215.49 & 6.00 & 8.65 & 0.87 & 0.39 \\ 
 \quad RL (B): & 1.0 & 0 & 1.0 & 205.41 & 5.77 & 7.76 & 0.87 & 0.40  \\ 
 \quad RL (C): & 1.0 & 1.0 & 0 & 230.95 & 6.55 & 8.68 & 0.88 & 0.37 \\ 
 \quad RL (D): & 1.0 & 1.0 & 1.0 & 211.67 & 5.91 & 8.58 & 0.86 & 0.40 \\ 
\bottomrule
\end{tabular}
\caption{Ablation study on reward design for ImageNet class-conditional $256\times 256$ image generation.
}
\label{tab:c2i-ablation-reward}
\end{table}

\noindent\textbf{Effect of Various Rewards.} 
As demonstrated in Figure~\ref{fig:overall-reward}, the reward exhibits continuous growth across training steps, consistent with our expectations. This indicates that the model successfully learns to generate patterns that enhance reward during the RL training process (See Figure~\ref{app:c2i-reward-details} and Figure~\ref{app:t2i-reward-details} in Appendix~\ref{app:reward-details} for details of individual reward). Notably, although reward exhibits continuous growth with increasing training steps, we observe that model performance deteriorates during later training phases, indicating significant reward hacking problems. Consequently, for evaluation purposes, we employ checkpoints at 100 and 300 steps for class-conditional and text-conditional image generation, respectively. 

The ablation study on reward resign for class-conditional image generation is presented in Table~\ref{tab:c2i-ablation-reward}. We observe that there is a clear trade-off between Precision (image quality) and Recall (diversity). For example, RL (A) achieves the highest quality but lowest diversity compared to the baseline LlamaGen-B and RL (D). The combination of conditional and image quality rewards without realism reward, \textit{i.e.}, RL (C), performs poorly, suggesting that realism is crucial for overall performance. Specifically, realism reward can maintain diversity as much as possible, but it will limit the improvement of image quality. We speculate that this is because VLM models use image data of various qualities during training and have higher requirements for generalizability. Although RL (D) with all three rewards does not excel in any single metric, it provides the most balanced performance across all evaluation criteria.

\begin{table}[h]
\centering
\begin{tabular}{@{}lccccccccccc@{}}
\toprule
\multirow{2}{*}{\bf Model} &  \multicolumn{3}{c}{\bf Reward Design} & \multirow{2}{*}{\bf PT$\uparrow$} & \multirow{2}{*}{\bf CL$\uparrow$} & \multirow{2}{*}{\bf AB$\uparrow$} & \multirow{2}{*}{\bf CT$\uparrow$} & \multirow{2}{*}{\bf SO$\uparrow$} & \multirow{2}{*}{\bf TO$\uparrow$} & \multirow{2}{*}{\bf Overall} \\ \cmidrule(lr){2-4}
& $\lambda_\text{C}$ & $\lambda_\text{I}$ & $\lambda_\text{R}$ & & & \\
\midrule
\our-XL &  & &  & 0.042 &  0.550 & 0.032 & 0.197 & 0.750 & 0.263 & 0.306  \\ 
 \quad RL (A): & 0 & 1.0 & 1.0 &  0.025 & 0.460 & 0.010 & 0.097 & 0.634 & 0.126 & 0.225  \\ 
 \quad RL (B): & 1.0 & 0 & 1.0 & 0.047 & 0.495 & 0.013 & 0.147 & 0.659 & 0.162 & 0.254 \\ 
 \quad RL (C): & 1.0 & 1.0 & 0 &  0.025 & 0.484 & 0.018 & 0.150 & 0.684 & 0.197 & 0.260 \\ 
 \quad RL (D): & 1.0 & 1.0 & 1.0 & 0.040 & 0.593 & 0.030 & 0.228 & 0.791 & 0.263 & 0.324 \\ 
\bottomrule
\end{tabular}
\caption{Ablation study on reward design for GenEval text-conditional $256\times 256$ image generation benchmark.
``PT'', ``CL'', ``AB'', ``CT'', ``SO'' and ``TO'' refer to \textit{position}, \textit{color}, \textit{attribute binding}, \textit{counting}, \textit{single object} and \textit{two object}, respectively.
}
\label{tab:t2i-ablation-reward-geneval}
\end{table}

In Table~\ref{tab:t2i-ablation-reward-geneval}, we show the ablation study on reward design for GenEval text-conditional image generation. When conditional reward is not used (\textit{i.e.}, RL (A)), there is a significant performance drop in the dimensions of \textit{position}, \textit{counting}, \textit{single object}, and \textit{two object}. This is because conditional reward provides constraints on whether the image content is consistent with the text, which directly affects the quality of visual semantic generation related to whether image content exists and quantity-related aspects. 
Similarly, when image quality reward and realism reward are respectively removed, the evaluation metrics also show quite significant drops.
When combining all three reward components yields superior results compared to using any sub-component in isolation. This suggests that high-quality text-to-image generation requires simultaneous optimization across multiple objectives.

\begin{table}[h]
\scriptsize
\centering
\begin{tabular}{@{}lcccccccccc@{}}
\toprule
\multirow{2}{*}{\bf Model} & \multicolumn{3}{c}{\bf Reward Design} & \multicolumn{2}{c}{\bf Alignment}  & \multicolumn{2}{c}{\textbf{Image Quality}} & \multicolumn{3}{c}{\textbf{Preference Score}} \\ \cmidrule(lr){2-4}\cmidrule(lr){5-6} \cmidrule(lr){7-8} \cmidrule(lr){9-11}
&  $\lambda_\text{C}$ & $\lambda_\text{I}$ & $\lambda_\text{R}$  & \textbf{CLIP Score} & \textbf{HPSv2} &  \textbf{Aesthetic} & \textbf{DeQA}  &\textbf{ImgRwd} & \textbf{PickScore} &  \textbf{UniRwd} \\
\midrule
\our-XL & & & & 0.245 & 0.153 & 4.701 & 0.546  & -1.784 & 0.737 & 0.271 \\
 \quad + RL (A): & 0 & 1.0 & 1.0 & 0.272 &  0.203 & 4.789 & 0.541  & -0.805 & 0.767 & 0.307  \\
 \quad + RL (B): & 1.0 & 0 & 1.0 &  0.274 & 0.206 & 4.770 & 0.536  & -0.726 & 0.769 & 0.314 \\
 \quad + RL (C): & 1.0 & 1.0 & 0 & 0.274 & 0.209 & 4.808 & 0.555  & -0.683 & 0.770 & 0.316  \\
 \quad + RL (D): & 1.0 & 1.0 & 1.0 & 0.274 & 0.208 & 4.808 & 0.551  &  -0.712 & 0.769 & 0.312 \\ 
\bottomrule
\end{tabular}
\caption{Ablation study on reward design for DrawBench~\cite{saharia2022photorealistic} text-to-image $256\times 256$ generation benchmark.
}
\label{tab:t2i-ablation-reward-drawbench}
\end{table}

We conduct an additional ablation study on the DrawBench~\cite{saharia2022photorealistic} benchmark, as shown in Table~\ref{tab:t2i-ablation-reward-drawbench}, examining how different reward design components affect text-to-image generation performance across text-image alignment, image quality, and human preference metrics. Surprisingly, RL (C) performs best overall, suggesting that realism reward may be less critical than conditional and quality rewards for this benchmark. However, RL (D) with all components still achieves strong, balanced performance across metrics. We observe several key insights: (1) All RL configurations significantly outperform the baseline, with substantial improvements in CLIP score, HPSv2 score and ImgRwd score. (2) Conditional reward appears most crucial for human preference, as configurations without it RL (A) shows the poorest preference scores. (3) Image quality reward strongly impacts aesthetic quality--RL (C) achieves the highest aesthetic and visual quality scores when combined with conditional reward. (4) Diminishing returns from realism reward--RL (D) does not substantially improve over RL (C), suggesting potential redundancy for this benchmark.

\begin{figure}[h]
\begin{minipage}{0.492\textwidth}
    \includegraphics[width=\textwidth]{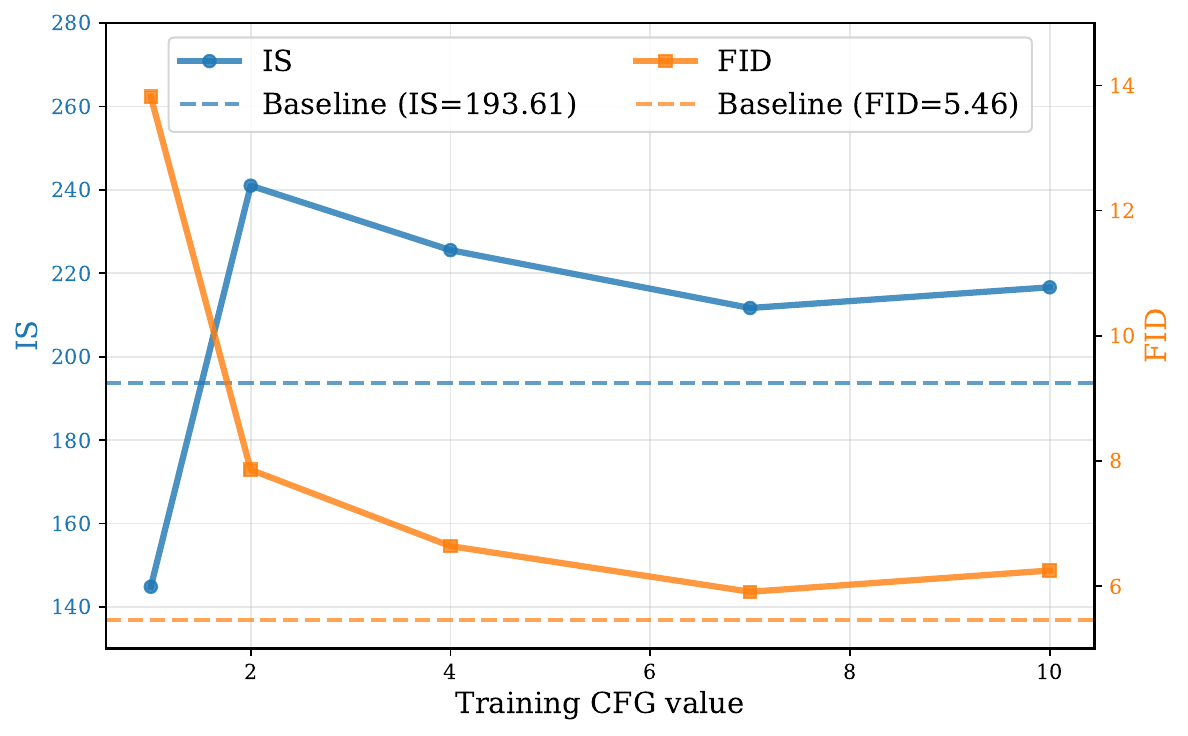}
\end{minipage}
 \hfill
\begin{minipage}{0.492\textwidth}
    \centering
    \includegraphics[width=\textwidth]{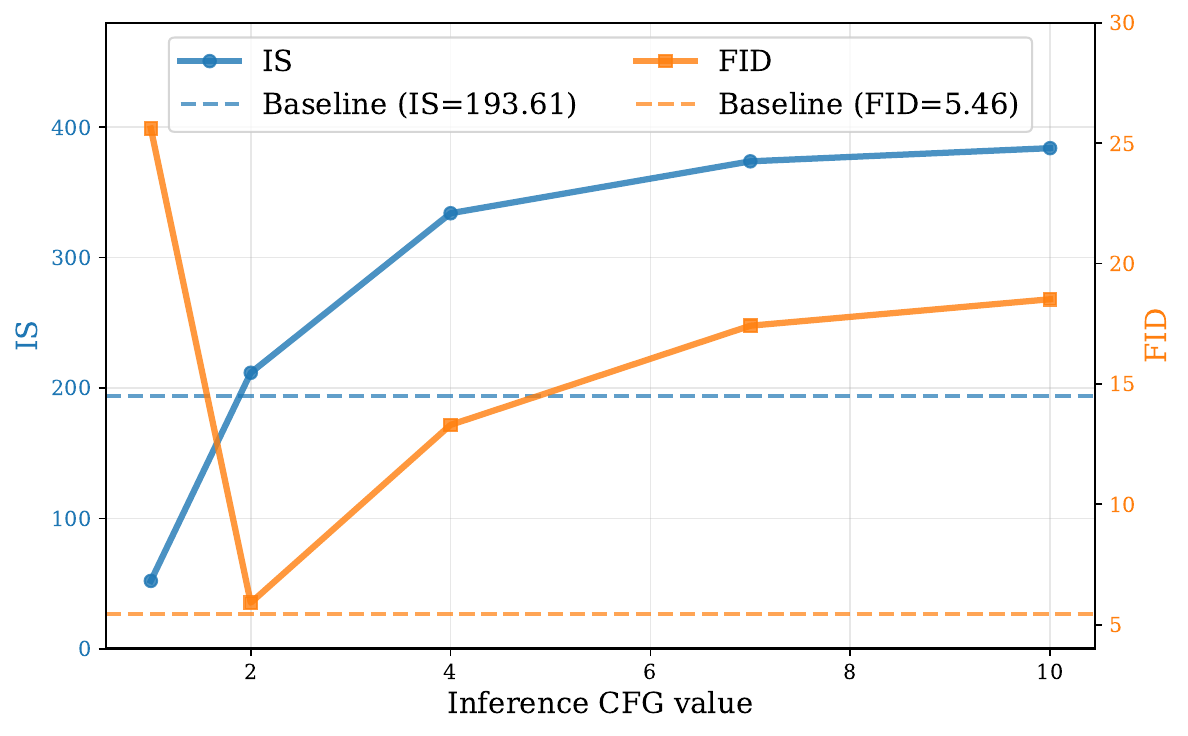}
\end{minipage}
\caption{The results of varying training and testing CFG weights on LlamaGen-B model for class-conditional $256\times 256$ image generation. 
(Left) We train multiple models with different CFG weights while maintaining a fixed inference CFG weight of s=2.0. (Right) We evaluate a model trained with CFG weight s=7.0 using various CFG weights during inference.}
\label{fig:cfg-ablation}
\end{figure}

\noindent\textbf{Effect of CFG Weight.}  
We analyze the effect of CFG weight in Figure~\ref{fig:cfg-ablation}. For training CFG variation (Figure~\ref{fig:cfg-ablation} (Left)), higher training CFG weights result in deteriorated performance on both metrics: increased FID performance (indicating better distribution alignment) and decreased IS performance (reflecting lower image quality and diversity). Moderate training CFG weights achieve optimal performance for the ``B'' size AR model, with $s=7.0$ providing the best balance.
For inference CFG variation (Figure~\ref{fig:cfg-ablation} (Right)), higher inference CFG weights consistently improve IS scores (reaching $\sim$390 at $s=10.0$). However, higher inference CFG weights also increase FID scores (degrading to $\sim$18 at $s=10.0$). There is a clear trade-off: better sample diversity (IS) comes at the cost of sample quality or distribution alignment (FID). Thus, training and inference CFG weights do not need to match models trained with one CFG weight can be effectively used with different inference CFG weights.
In practice, training with CFG $s=2.0\sim8.0$, then adjust inference CFG based on whether quality (lower CFG) or diversity (higher CFG) is more important for the specific application. Finally, we extend our CFG weight analysis to larger-scale models and observe that the optimal CFG weight ranges for training and inference also vary with model size, aligning with findings reported in LlamaGen.

\begin{figure}[h]
\centering
\includegraphics[width=\linewidth]{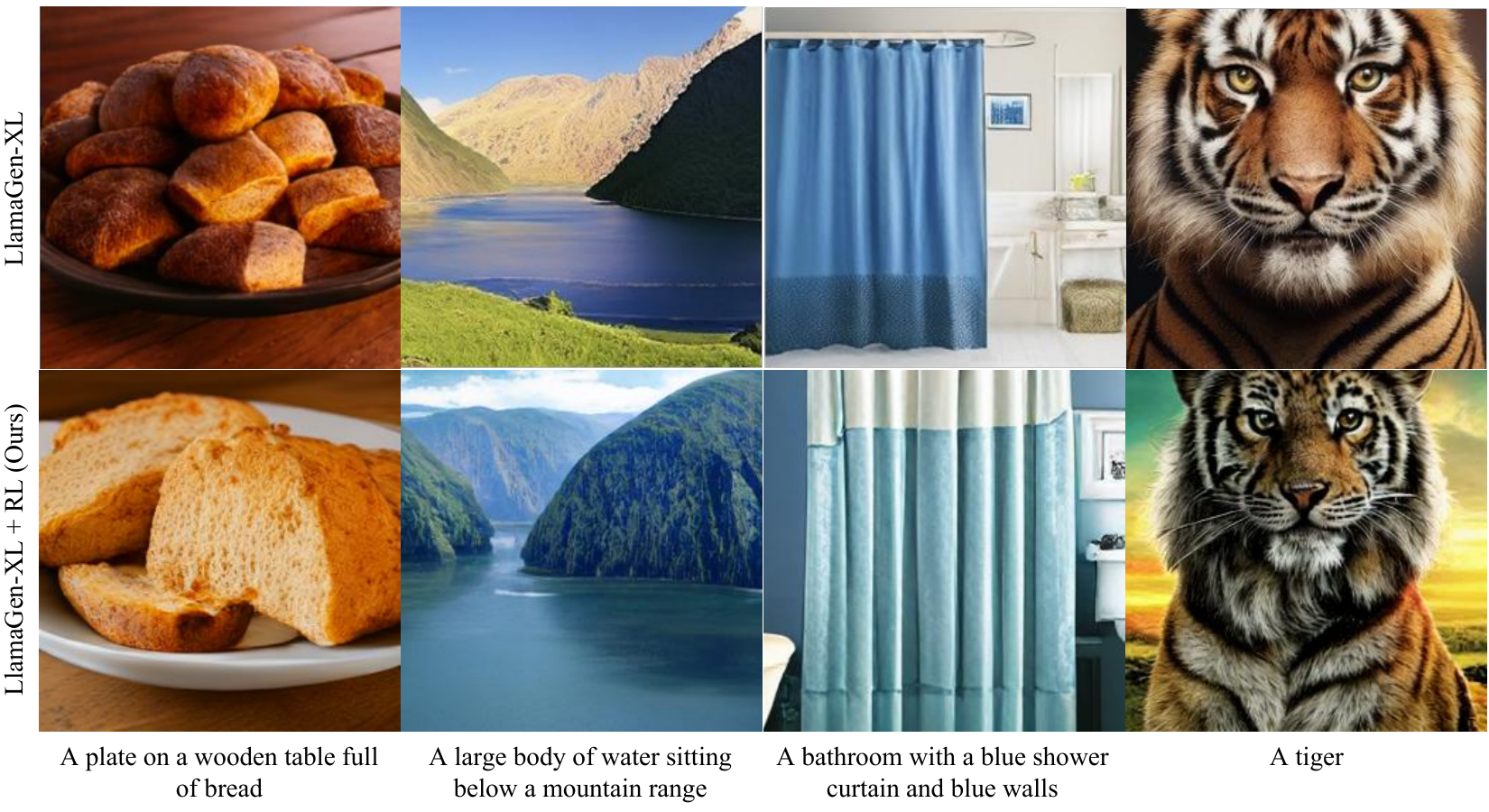}
\caption{Visual comparisons bwteen LlamaGen-XL model and our RL trained model on text-conditional $256\times 256$ image generation task.}
\label{fig:t2i-vis-comparison}
\end{figure}

\subsection{Visual Comparisons} We show several visual comparisons between the original LlamaGen-XL~\cite{sun2024autoregressive} model and our RL trained model on text-conditional image generation in Figure~\ref{fig:t2i-vis-comparison}. Visual comparison reveals that RL-trained models demonstrate superior image quality, especially regarding realism, with results that better align with human perceptual preferences. Close examination of the third bathroom image set (\textit{i.e.}, ``A bathroom with a blue shower curtain and blue wall'') reveals that while the baseline model overlooks the textual semantics of ``blue walls'' and produces only a blue curtain, our RL model demonstrates comprehensive understanding of the text prompt by generating both blue curtain and walls, exhibiting superior text-visual alignment.
We refer readers to Appendix~\ref{app:visual-results} for more visual results for text-conditional image generation.

\section{Conclusion, Limitation, and Future Work}
\label{sec:conclusion}
In this work, we explore a crucial question: whether reinforcement learning (RL) techniques that have achieved breakthrough success in LLMs can enhance autoregressive (AR) image generation models. By introducing the GRPO~\cite{shao2024deepseekmath} algorithm into the LlamaGen~\cite{sun2024autoregressive} framework and designing multi-dimensional rewards—including conditional rewards for semantic alignment (what and where), image quality rewards for visual coherence and clarity, and realism rewards for natural appearance. We conducted RL training on both class-conditional (\textit{class-to-image}) and text-conditional (\textit{text-to-image}) image generation tasks.
Experimental results demonstrate that RL training in the \textit{class-to-image} scenario contributes to higher quality and accuracy in matching the visual semantics of corresponding categories, but slightly affects the overall data distribution shift and category coverage or diversity. In the \textit{text-to-image} scenario, RL training helps improve performance across multiple dimensions, including image-text alignment, image quality, and human preference alignment.

\noindent\textbf{Limitations.} LlamaGen provides a very comprehensive series of foundation models of different sizes as pre-trained models, which is highly advantageous for conducting various RL post-training experiments. However, due to the baseline model capabilities, there is still a considerable gap compared to SD-based models~\cite{liu2025flow,xue2025dancegrpo}. During our exploration process, we realize that compared to SD-based RL methods, there remains a substantial gap. Additionally, in the \textit{text-to-image} scenario, we do not conduct task-oriented reward integration experiments, such as Flow-GRPO~\cite{liu2025flow} employs OCR and GenEval rewards, which limits the evaluation performance on corresponding tasks. This also indicates that this direction still holds tremendous potential for exploration.

\noindent\textbf{Future Work.} Recent VAR~\cite{tian2024visual} shows a promising direction to model images in a coarse-to-fine ``next-scale prediction'' manner. Unlike the long chain-of-thought (CoT) reasoning in recent LLMs, VAR's next-scale prediction is a well-structured generation paradigm. 
We are very interested in reasoning approaches in the visual token space that more closely resemble LLM long CoT reasoning, and we will explore and attempt this in our future work. Additionally, we observe a reduction in model policy entropy after RL training, leading to more deterministic sampling that harms generation diversity--an area deserving further investigation and enhancement.

\bibliographystyle{unsrtnat}
\bibliography{main}

\clearpage
\appendix

\section{Experimental Resources}
\label{app:exp-resource}

For the LlamaGen~\cite{sun2024autoregressive} series models, we use official source codes~\footnote{\url{https://github.com/FoundationVision/LlamaGen}} and corresponding released model weights.  For the GRPO algorithm, we employ the implementation in HuggingFace TRL~\footnote{\url{https://github.com/huggingface/trl}}. We train all of our models with 8 NVIDIA H800 GPUs. 

\section{More Visual Results}
\label{app:visual-results}

\begin{figure}[ht]
\vspace{-1em}
\centering
\includegraphics[width=0.82\linewidth]{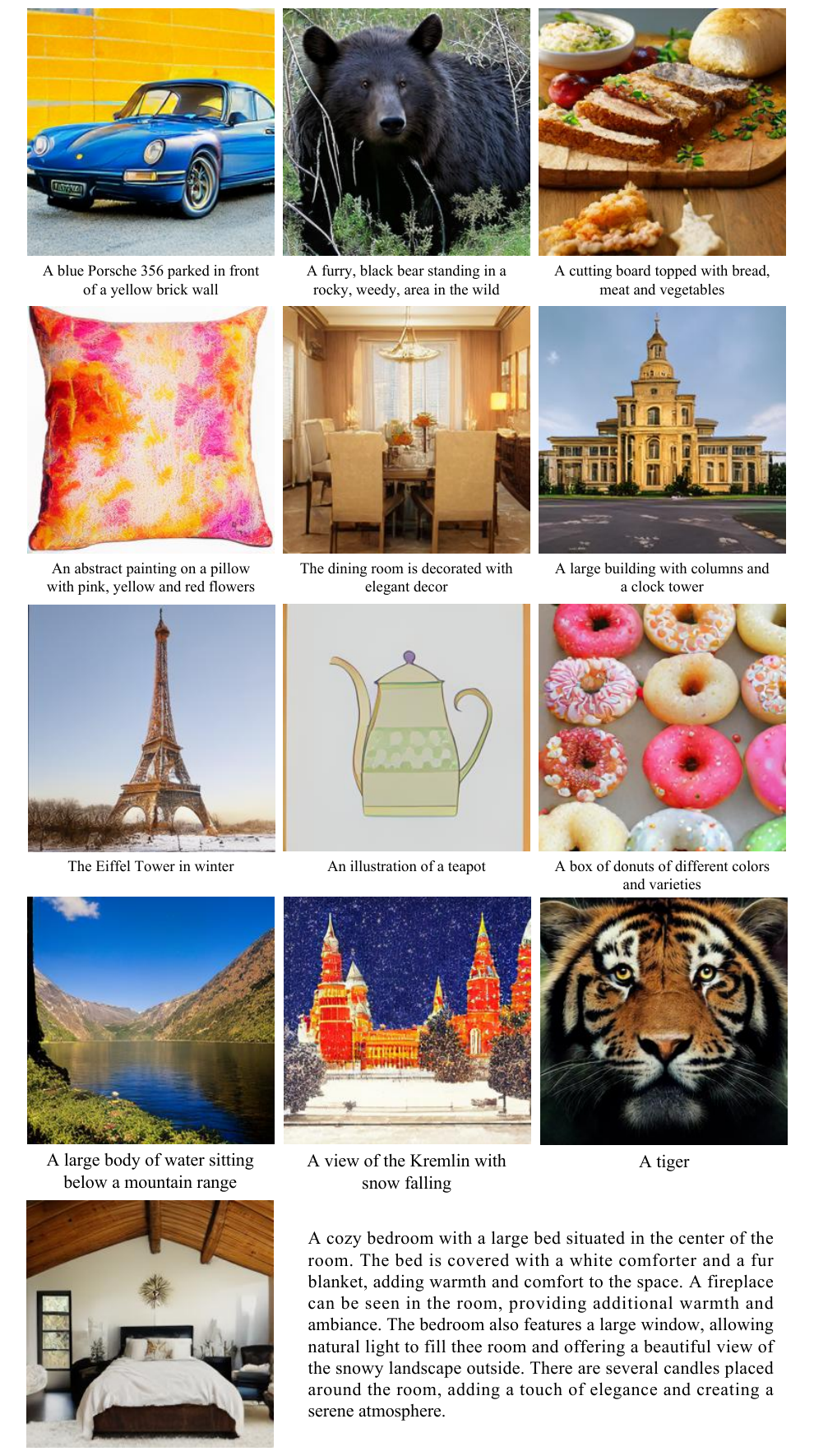}
\caption{Visual results of text-conditional $256\times 256$ image generation with our RL trained model.}
\label{fig:t2i-app}
\end{figure}

In Figure~\ref{fig:c2i_014}$\sim$ Figure~\ref{fig:c2i_884}, and Figure~\ref{fig:t2i-app}, we show more visual results on class-conditional and text-conditional image generation respectively.  

\section{VLM Prompts}
\label{app:vlm-prompts}

We employ the Qwen2.5-VL-3B-Instruct~\cite{qwen2.5-VL} model as a judger to evaluate generated image realism along three aspects:
\begin{itemize}
    \item \textit{Fake Identification} to analyze images to determine whether they are artificially generated (by AI models) or authentic/real photographs.
    \item \textit{Label Recognition} to evaluate whether the generated image correctly depicts an object that matches the specified category or text condition mentioned in the input prompt.
    \item \textit{Weird Detection} to inspect images to identify strange or unusual features.
\end{itemize}

Our prompt is shown in Figure~\ref{app:vlm-prompt}, where we instruct the VLM to assess all three aspects in a single evaluation.

\section{Reward Details in the Training Process}
\label{app:reward-details}

\begin{figure}[t]
\begin{minipage}{0.49\textwidth}
    \includegraphics[width=\textwidth]{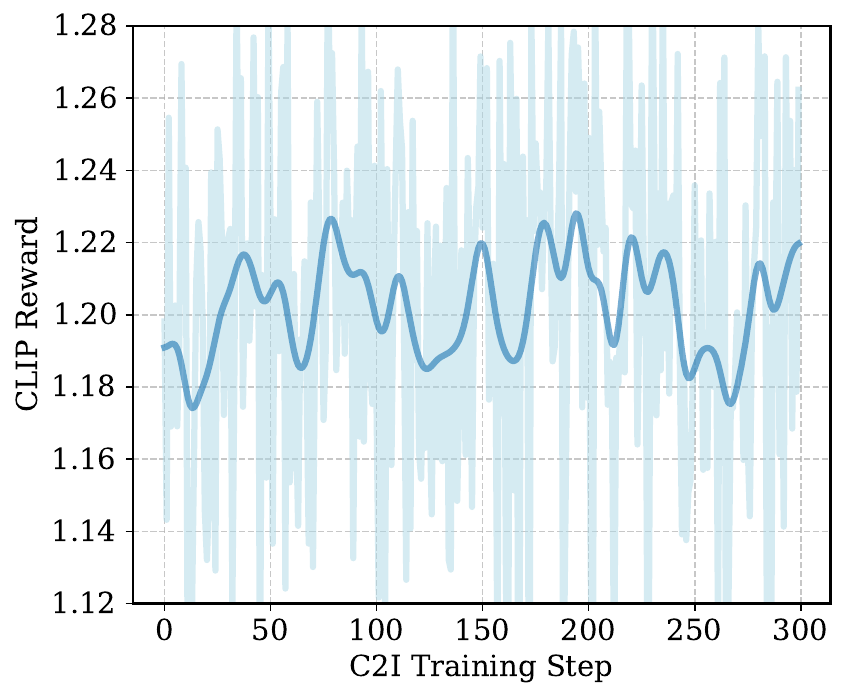}
\end{minipage}
 \hfill
\begin{minipage}{0.49\textwidth}
    \centering
    \includegraphics[width=\textwidth]{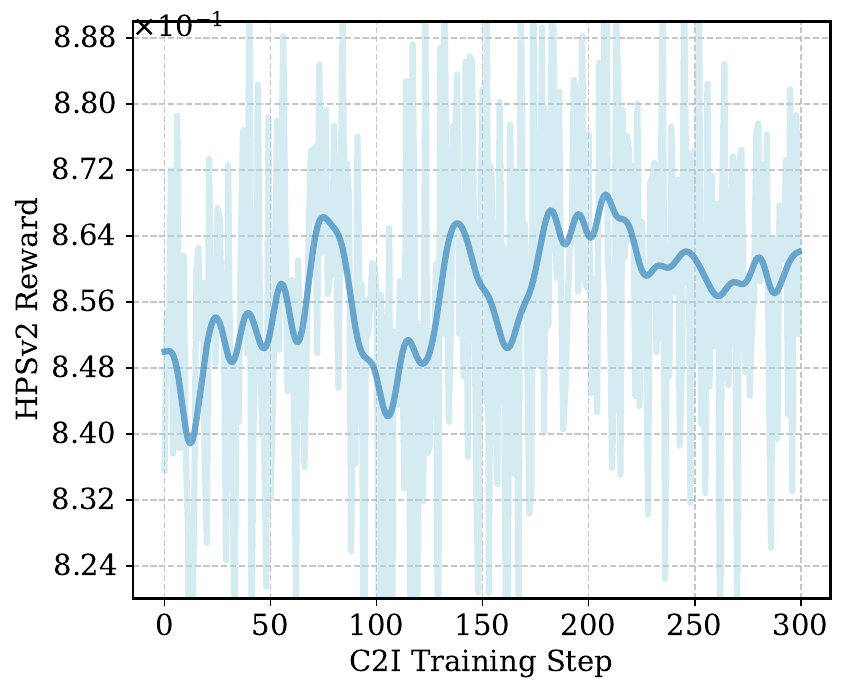}
\end{minipage}
 \hfill
\begin{minipage}{0.49\textwidth}
    \centering
    \includegraphics[width=\textwidth]{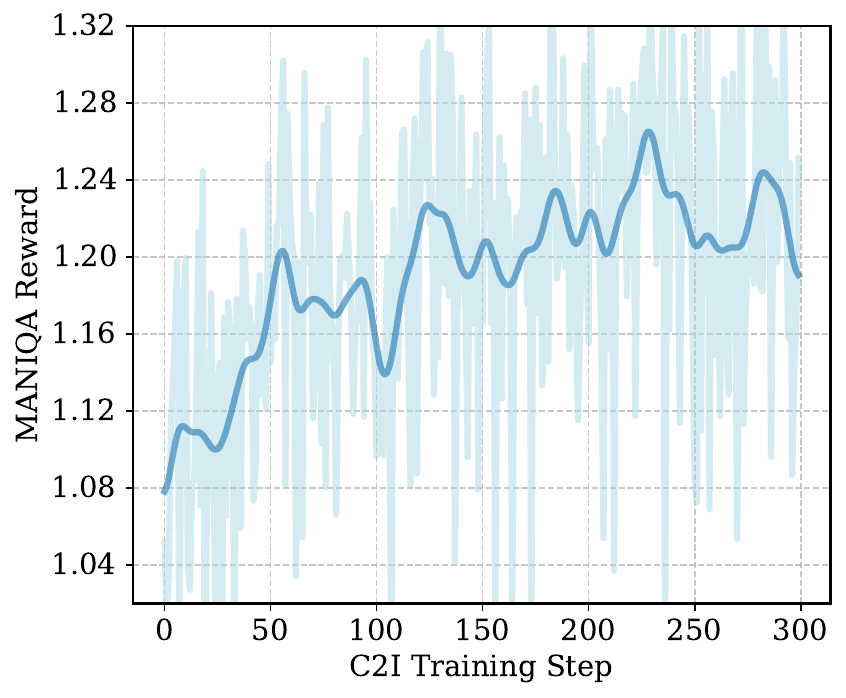}
\end{minipage}
 \hfill
\begin{minipage}{0.49\textwidth}
    \centering
    \includegraphics[width=\textwidth]{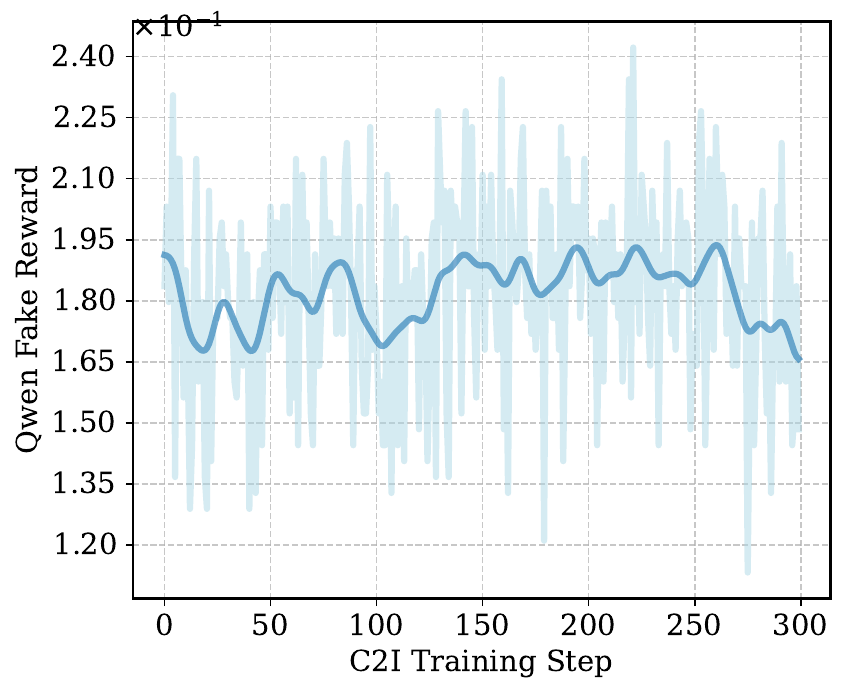}
\end{minipage}
 \hfill
\begin{minipage}{0.49\textwidth}
    \centering
    \includegraphics[width=\textwidth]{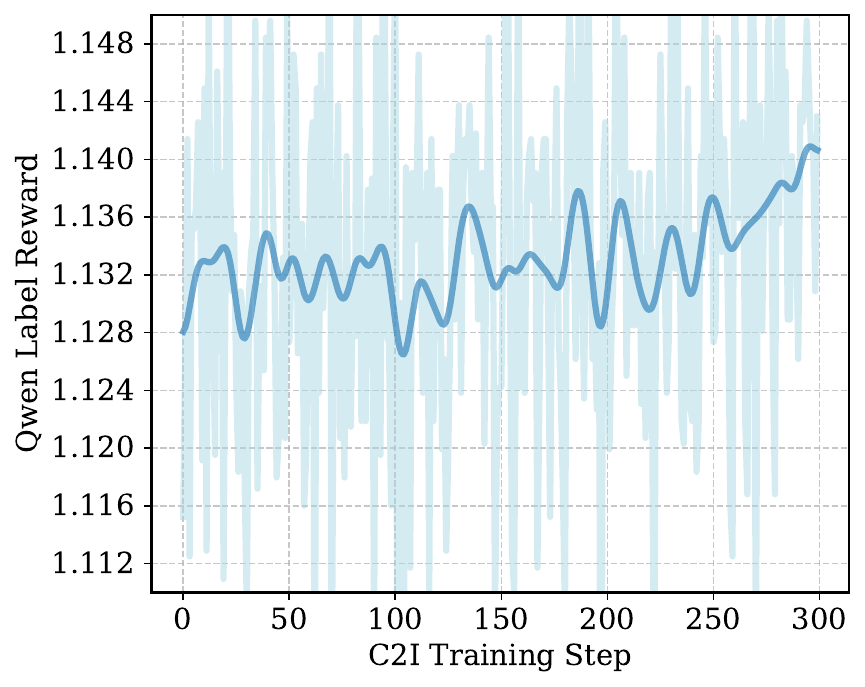}
\end{minipage}
 \hfill
\begin{minipage}{0.49\textwidth}
    \centering
    \includegraphics[width=\textwidth]{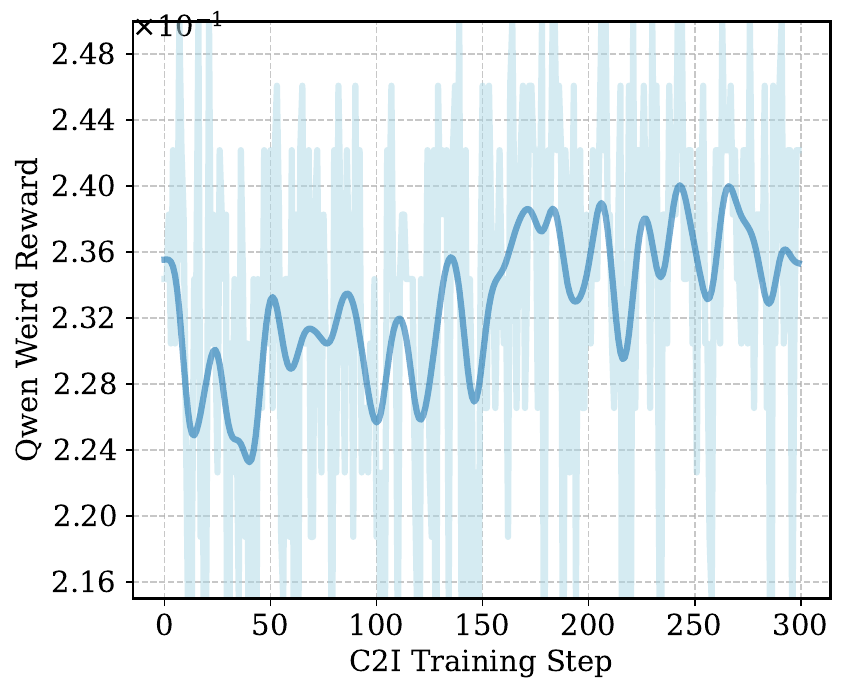}
\end{minipage}
\caption{Detailed reward curves during the RL training for class-conditional image generation models.}
\label{app:c2i-reward-details}
\end{figure}

\begin{figure}[t]
\begin{minipage}{0.49\textwidth}
    \includegraphics[width=\textwidth]{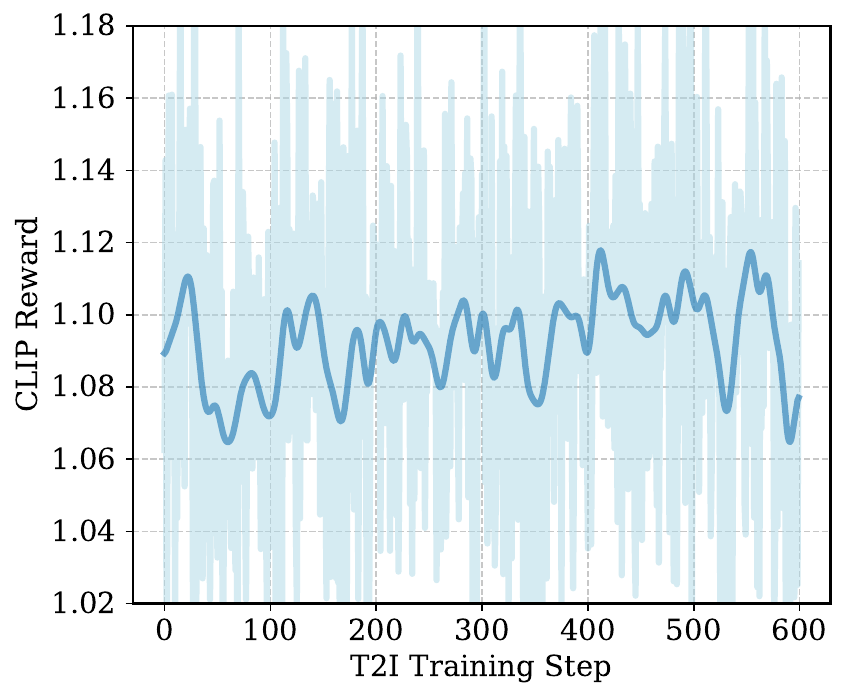}
\end{minipage}
 \hfill
\begin{minipage}{0.49\textwidth}
    \centering
    \includegraphics[width=\textwidth]{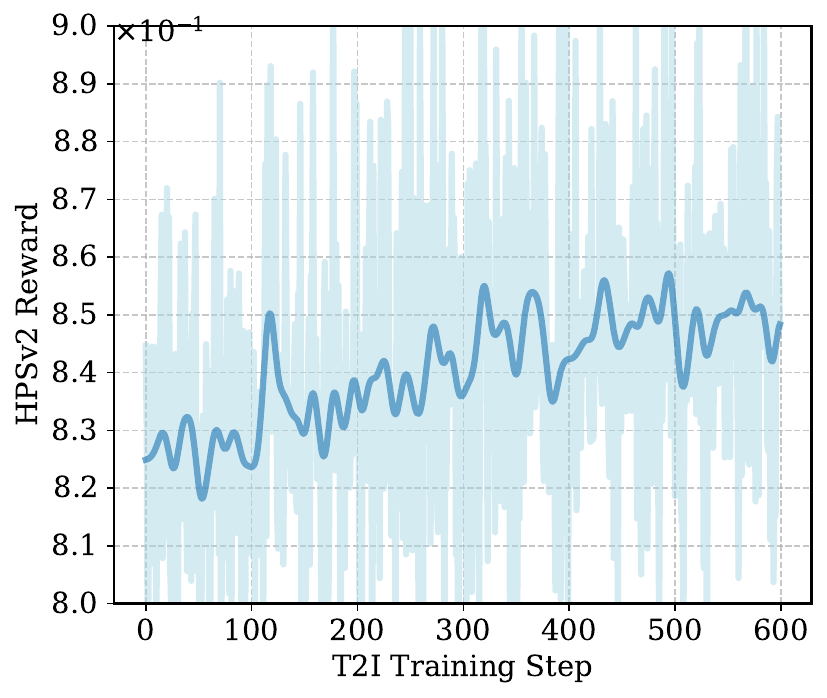}
\end{minipage}
 \hfill
\begin{minipage}{0.49\textwidth}
    \centering
    \includegraphics[width=\textwidth]{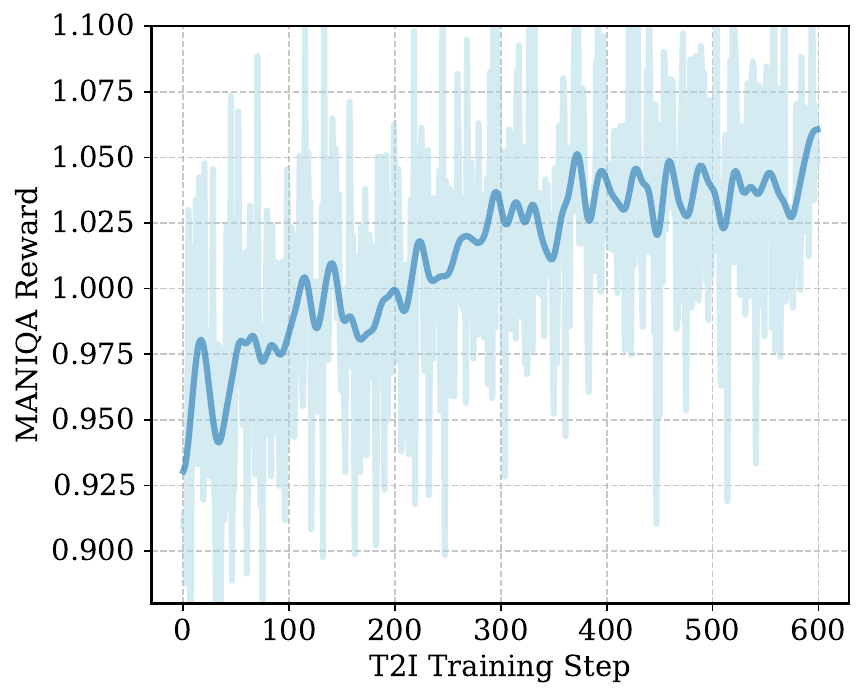}
\end{minipage}
 \hfill
\begin{minipage}{0.49\textwidth}
    \centering
    \includegraphics[width=\textwidth]{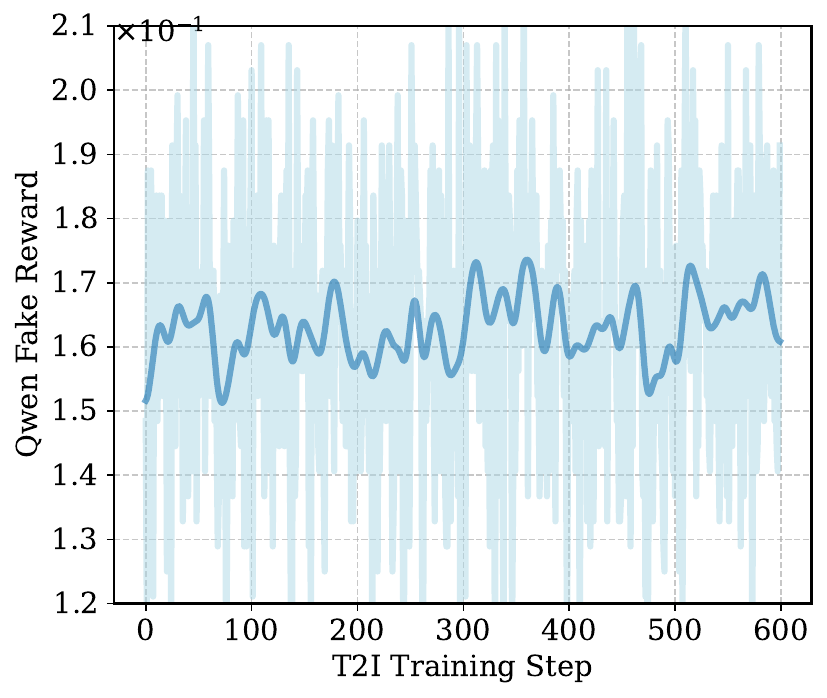}
\end{minipage}
 \hfill
\begin{minipage}{0.49\textwidth}
    \centering
    \includegraphics[width=\textwidth]{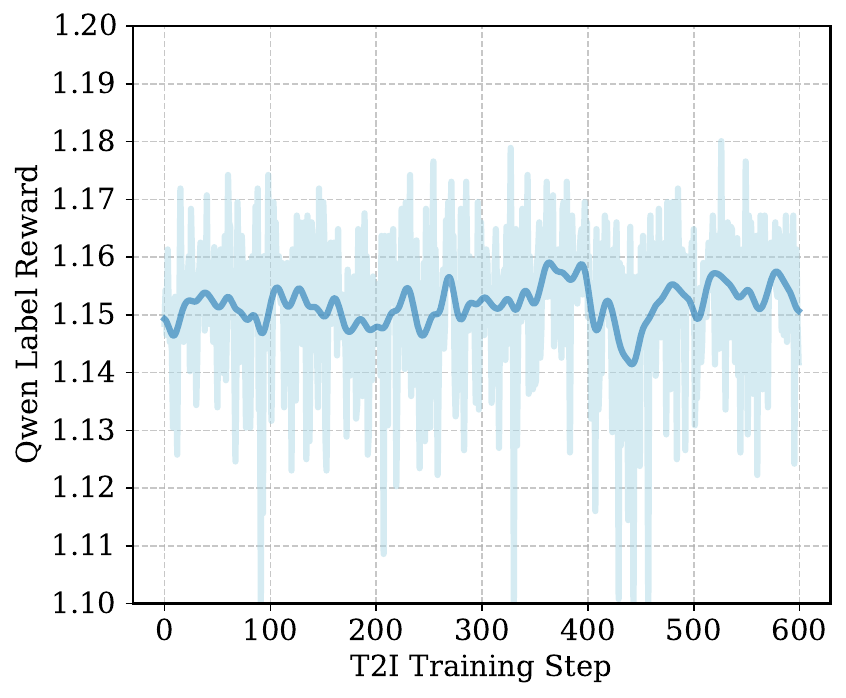}
\end{minipage}
 \hfill
\begin{minipage}{0.49\textwidth}
    \centering
    \includegraphics[width=\textwidth]{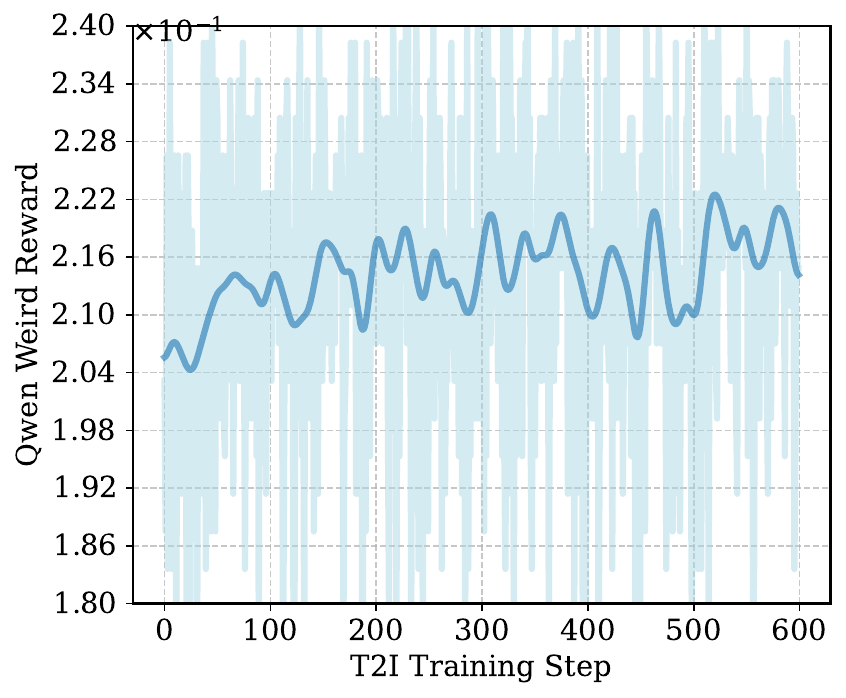}
\end{minipage}
\caption{Detailed reward curves during the RL training for text-conditional image generation models.}
\label{app:t2i-reward-details}
\end{figure}

During training, to balance the impacts of different rewards, the outputs of reward models are scaled to values roughly from 1 to 2. Specifically, CLIP score and HPSv2 score are multiplied by 5; MANIQA score is multiplied by 2; Qwen rewards are multiplied by 0.25. Also, to mitigate reward hacking of CLIP score and HPSv2 score, we introduce the quantized version of these two models, with continuous values being mapped to three discrete values~(0.5,1,1.5). The quantized rewards are used along with the original CLIP and HPSv2 rewards during training. As shown in Figure~\ref{app:c2i-reward-details} and Figure~\ref{app:t2i-reward-details}, we present the detailed reward curves for both class-conditional and text-conditional image generation during the RL training. In these two scenarios, we find that HPSv2 reward, MANIQA reward, and Qwen Weird reward show clear increasing trends, while CLIP reward, Qwen Fake reward, and Qwen Label reward remain relatively stable throughout training.

\begin{minipage}[ht]{0.49\textwidth}
    \begin{figure}[H]
    \centering
    \includegraphics[width=0.85\textwidth]{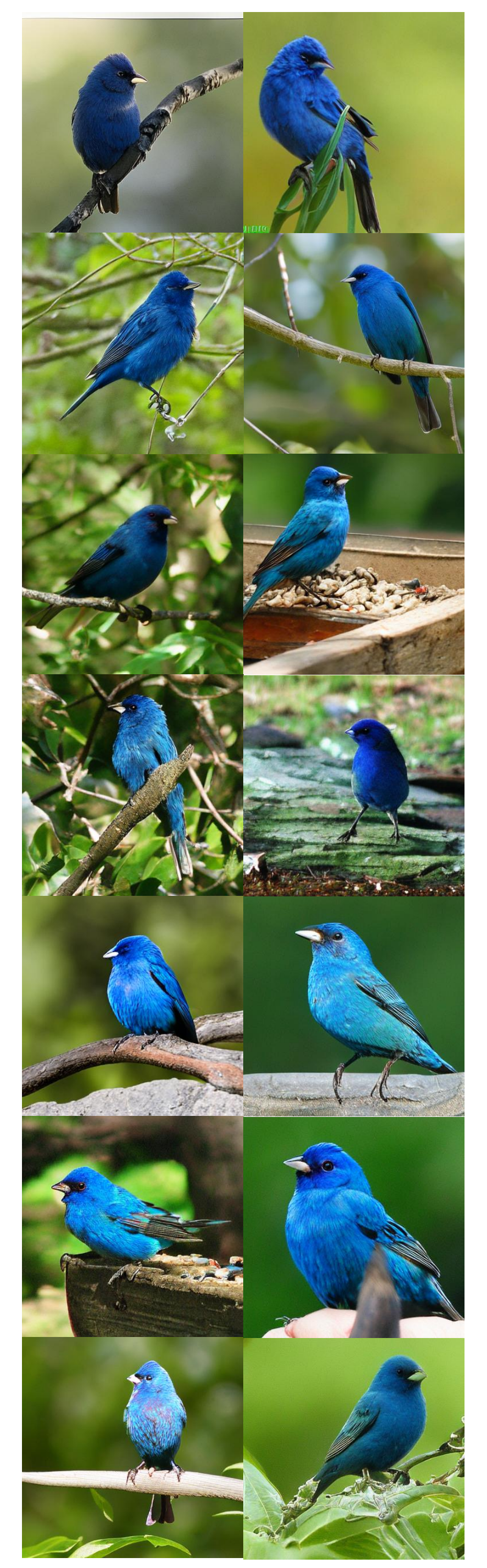}
    \caption{384$\times$384 \our-XL samples. \\ Classifier-free guidance scale $s=2.0$ \\
    Class label = ``indigo bunting'' (014)}
    \label{fig:c2i_014}
    \end{figure}
\end{minipage}
\hfill
\begin{minipage}[ht]{0.49\textwidth}
    \begin{figure}[H]
    \centering
    \includegraphics[width=0.85\textwidth]{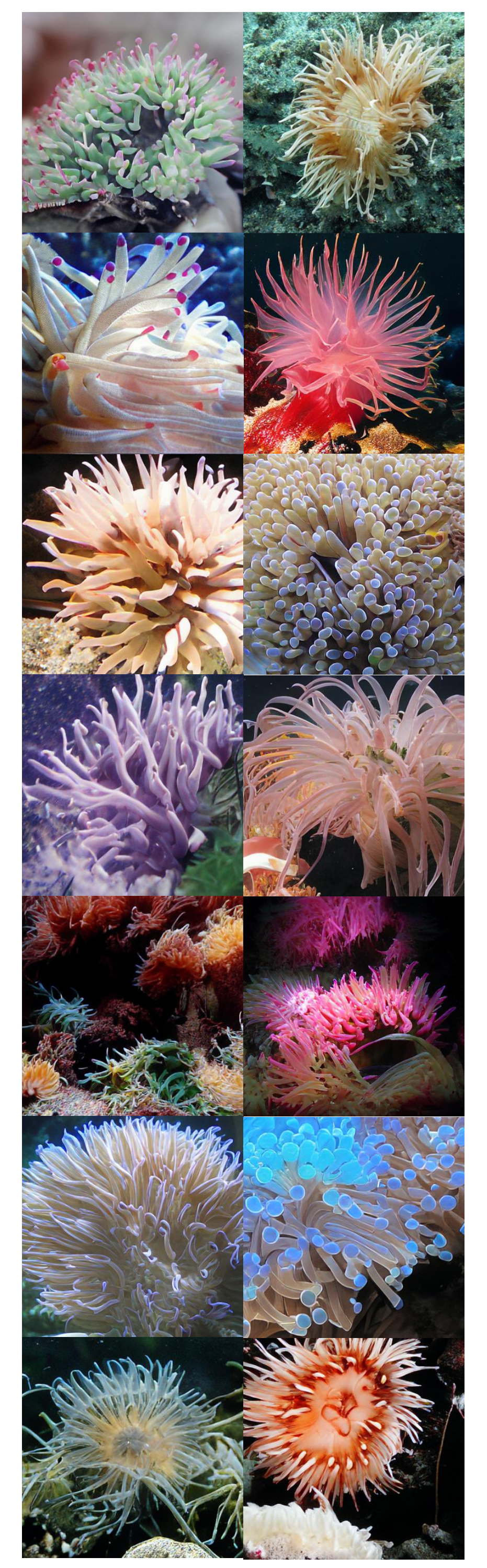}
    \caption{384$\times$384 \our-XL samples. \\
    Classifier-free guidance scale $s=2.0$ \\
    Class label = ``sea anemone'' (108)}
    \label{fig:c2i_108}
    \end{figure}
\end{minipage}

\begin{minipage}[ht]{0.49\textwidth}
    \begin{figure}[H]
    \centering
    \includegraphics[width=0.85\textwidth]{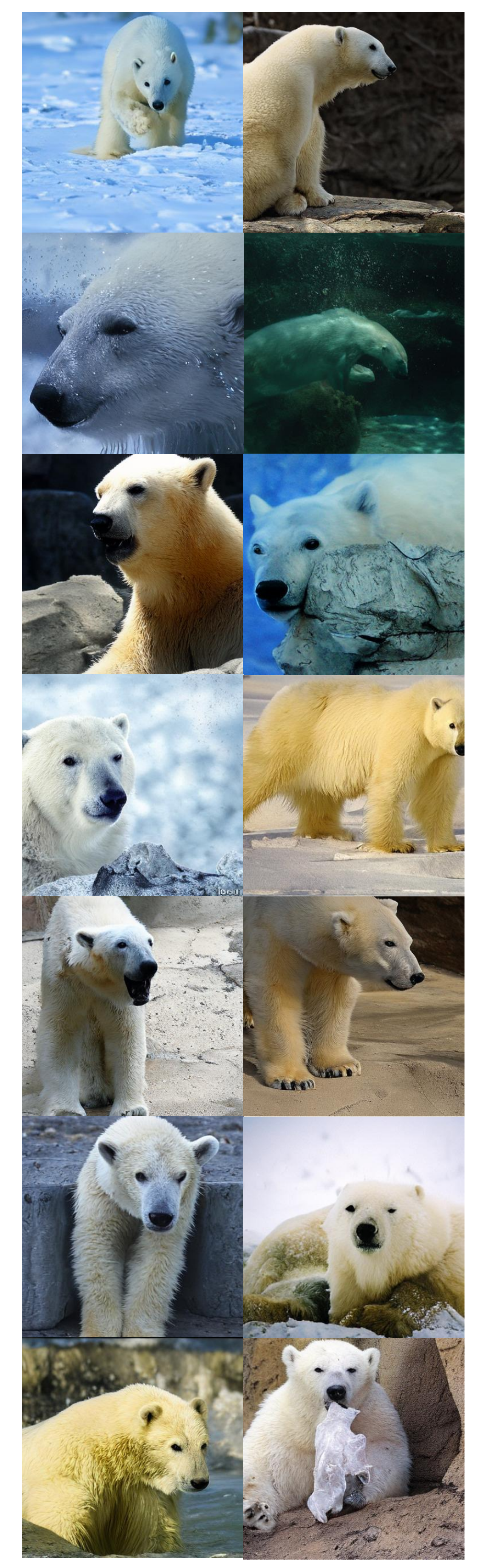}
    \caption{384$\times$384 \our-XL samples. \\ Classifier-free guidance scale $s=2.0$ \\
    Class label = ``ice bear'' (296)}
    \label{fig:c2i_296}
    \end{figure}
\end{minipage}
\hfill
\begin{minipage}[ht]{0.49\textwidth}
    \begin{figure}[H]
    \centering
    \includegraphics[width=0.85\textwidth]{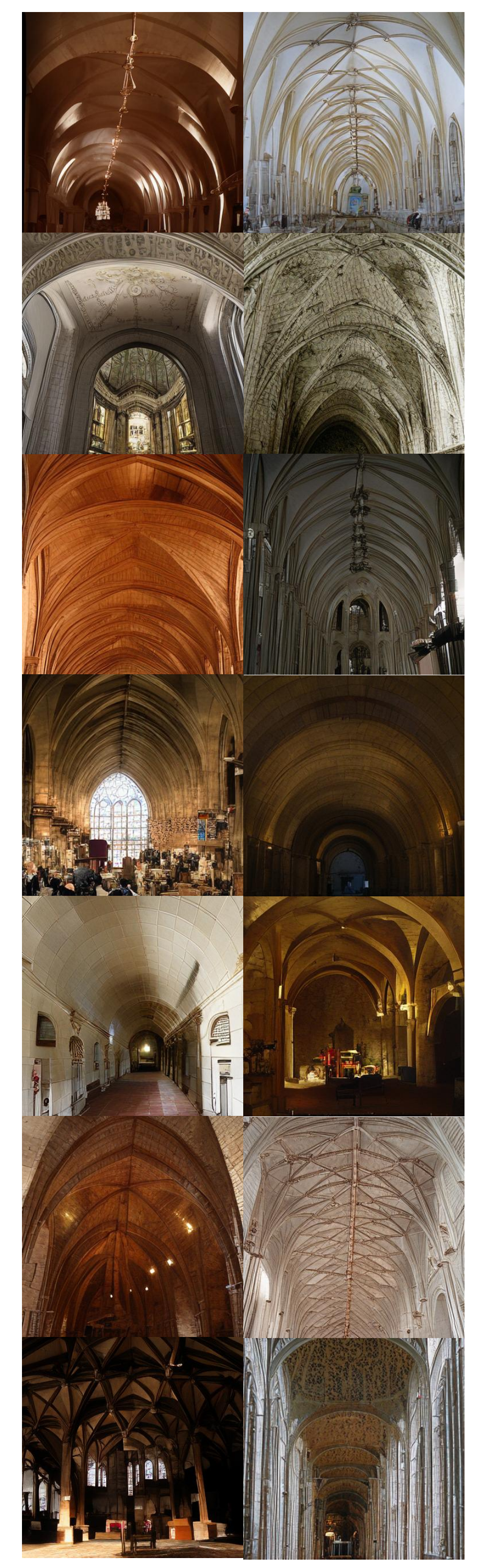}
    \caption{384$\times$384 \our-XL samples. \\
    Classifier-free guidance scale $s=2.0$ \\
    Class label = ``vault'' (884)}
    \label{fig:c2i_884}
    \end{figure}
\end{minipage}

\begin{figure}[t]
    \centering
    \begin{center}
    \begin{promptbox}[Prompt for VLM-as-a-judge Method]
You are given a text prompt used to generate image: "\{\}" \newline
Below is one generated image: <image>\newline

1. Describe the image thoroughly (objects, colors, layout, etc.), do not be affected by the prompt.\newline
2. Score the generation quality from following aspects:\newline
- Is the object in the generated image satisfying the category according to the prompt? (0-1 score) \newline
- Is the generated image completed without missing parts? (0-1 score) \newline
- Does the content of generated image look real and reasonable? (0-1 score) \newline
- Is the generated image clear, bright and in high quality? (0-1 score) \newline
- Is the generated image free of any noises, defects and artifacts? (0-1 score) \newline

Your response should in a **JSON** format following rules below:\newline
1. The final response should have three keys: "description", "score", "explanation". \newline
2. The total score of the image should be store in "score" key, you may use float numbers for the scores. The value of total score should be between 0 and 5.\newline
3. The reasoning and scoring process could be write in "explanation" for you to explain your score.
Following is a exmaple of response:\newline

\text{`}\text{`}\text{`}json\newline
\{ \newline
"description": "The image shows a brown dog with one eye and two ears. its mouth is open and the its tongue is sticking out.",\newline
\quad \quad "score": 3.5,\newline
"explanation": "The image shows a brown dog matches the prompt \text{`}a photo of a dog\text{'} (1 score). The image is complete and all part can be seen (1 score). The dog in the photo looks unreal because it has only one eye and the eye position is not reasonable (0 score). The dog's face is clear but its body and background is blurred (0.5 score). The image do not have any noises, defects and artifacts (1 score). So the total score of this image is 1+1+0+0.5+1=3.5 .\text{"}\newline
\}\newline
\text{'}\text{'}\text{'} \newline

<image>\newline
You need to inspect the image carefully and give a score according to the question below:\newline
- Is the image an AI-generated image? If yes, give 0 score, otherwise give 1 score.\newline
The answer should be in a JSON format as follows:\newline
\text{`}\text{`}\text{`}json\newline
\{"score": 0\}\newline
\text{'}\text{'}\text{'}\newline

<image>\newline
You need to inspect the image carefully and give a score according to the question below: \newline
- Is there any strange feature in the image? If yes, give 0 score, otherwise give 1 score.\newline
The answer should be in a JSON format as follows:\newline
\text{`}\text{`}\text{`}json\newline
\{"score": 0\}\newline
\text{'}\text{'}\text{'}\newline
    \end{promptbox}
    \end{center}
    \caption{Prompt for VLM-as-a-judge method to analyze the pair of text and generated image.}
    \label{app:vlm-prompt}
\end{figure}

\end{CJK*}
\end{document}